%% file: main.tex
\pdfoutput=1
\documentclass{article}

    \PassOptionsToPackage{numbers, sort, compress}{natbib}

\usepackage[preprint]{neurips_2025}
\usepackage[font=small]{caption}

\usepackage[utf8]{inputenc} 
\usepackage[T1]{fontenc}    
\usepackage{url}            
\usepackage{booktabs}       
\usepackage{amsfonts}       
\usepackage{nicefrac}       
\usepackage{microtype}      
\usepackage{xcolor}         

\usepackage{amsmath}
\usepackage{amssymb}
\definecolor{linkblue}{HTML}{6870ae}
\usepackage[pagebackref=true,colorlinks,citecolor=brown]{hyperref}
\hypersetup{
  urlcolor  = linkblue
}
\usepackage{multirow}
\usepackage{makecell}
\usepackage{colortbl}       
\usepackage{wrapfig}
\usepackage[inkscape=false]{svg} 
\usepackage{cleveref}
\usepackage{caption}
\usepackage{fontawesome5}

\usepackage{graphicx}
\usepackage{float}
\floatstyle{plaintop}
\restylefloat{table}
\usepackage[export]{adjustbox}

\usepackage{amsthm}
\usepackage{subcaption}
\usepackage{etoolbox}

\usepackage{pifont}

\usepackage{tikz}
\usetikzlibrary{arrows.meta}

\usepackage{fontawesome5}
\newcommand{\blfootnote}[1]{%
  \begingroup
    \renewcommand\thefootnote{}
    \footnote{#1}%
    \addtocounter{footnote}{-1}
  \endgroup}

\usepackage[ruled,vlined,linesnumbered,noend]{algorithm2e}
\SetKwInput{Inputs}{\textbf{Inputs}}
\SetKwInput{Hparams}{\textbf{Hparams}}

\title{Explorative Modeling: Unlocking a Third Pretraining Axis and End-to-End Generation}

\usepackage{epigraph}
\usepackage[most]{tcolorbox}
\definecolor{TakeBlue}{RGB}{28,81,214} 
\definecolor{TakeBack}{RGB}{238,241,255} 
\newtcolorbox{takeaway}{
  colback=TakeBack,
  colframe=TakeBlue,
  boxrule=0.9pt,
  arc=3mm,                
  left=8pt,right=8pt,top=6pt,bottom=6pt,
}
\definecolor{DefTeal}{RGB}{20,130,140}   
\definecolor{DefBack}{RGB}{235,248,248}  
\newtcolorbox{definitionbox}{
  colback=DefBack,
  colframe=DefTeal,
  boxrule=0.9pt,
  arc=3mm,
  left=8pt,right=8pt,top=6pt,bottom=6pt,
}

\author{%
      \textbf{Alexi Gladstone}\textsuperscript{1}, \textbf{Heng Ji}\textsuperscript{1}, \textbf{Yilun Du}\textsuperscript{2}\\
      \textsuperscript{1}UIUC \quad \textsuperscript{2}Harvard \\[0.6em]
      \textbf{\href{https://explorative-modeling.github.io}
         {\faGlobe\enspace{explorative-modeling.github.io}}
        \quad
      \href{https://github.com/alexiglad/XM}
         {\faGithub\enspace{github.com/alexiglad/XM}}}
}

\NewDocumentCommand{\heng}
{ mO{} }{\textcolor{red}{\textsuperscript{\textit{Heng}}\textsf{\textbf{\small[#1]}}}}

\begin{document}

\blfootnote{Correspondence to Alexi Gladstone:  \href{mailto:alexigladstone@gmail.com}{\faEnvelope\enspace{alexigladstone@gmail.com}}. Work done while supported as a Flapping Airplanes Fellow.}

\maketitle

\input{main_sec/0_abstract}
\input{main_sec/1_intro}

\input{main_sec/2_background}
\input{main_sec/3_approach}
\input{main_sec/4_experimentation}
\input{main_sec/6_discussion}

\input{main_sec/7_broader_impact_future_work}
\input{main_sec/8_conclusion_limitations}

\input{main_sec/9_acknowledgements}

\clearpage
\newpage
{
    \small
    \bibliographystyle{unsrtnat}
    \bibliography{references}
}

\clearpage
\newpage
\appendix

\renewcommand{\thesection}{\Alph{section}}          
\renewcommand{\thefigure}{\Alph{section}\arabic{figure}} 
\renewcommand{\thetable}{\Alph{section}\arabic{table}}   

\setcounter{section}{0}
\counterwithin{figure}{section}   
\counterwithin{table}{section}

\clearpage
\input{supp_sec/A_additional_experiments}

\input{supp_sec/B_Additional_Intuition}

\input{supp_sec/C_Approach_Details}
\input{supp_sec/D_Experimental_Details}
\input{supp_sec/E_related_works}
\input{supp_sec/F_Additional_Theory}
\input{supp_sec/G_FAQ}

\end{document}

%% file: main_sec/0_abstract.tex
\begin{abstract}
\label{sec:abstract}

The deep learning revolution, kicked off by AlexNet, taught us that end-to-end training beats decomposing a problem into hand-designed stages.
Generative modeling, however, has remained the exception---despite generative models being remarkably capable, they are still not trained end-to-end.\footnote{The term end-to-end generative modeling is often used loosely. We provide a stricter definition in Section~\ref{sec:background}---simply put, sampling during training should be the same as sampling during inference.} This is because, at its core,
\textit{generative modeling is about handling multimodal distributions},\footnote{By \textit{multimodal} we mean a probability distribution with many modes (distinct peaks), not data of different modalities such as text and images.} and existing scalable approaches handle this multimodality the same way, by \textit{factoring the generation procedure}, which prevents end-to-end generation.
In this work, we introduce \textbf{Explorative Modeling}, a new paradigm that instead factors the training loop, exploring $K$ candidate matches between model generations and data, and training on the best, so predictions commit to modes rather than blurring them.
We find Explorative Models (\textbf{XMs}) useful in two settings. First, increasing exploration adds a \textbf{third pretraining axis} \textit{beyond parameters and data} for existing generative models---where scaling exploration monotonically improves performance across both continuous and discrete domains (images, video, and language).
Notably, gains from exploration \textbf{increase with scale}, climbing from $7\%$ to $36\%$ as data scales and from $13\%$ to $23\%$ as models grow, with efficiency gains more than doubling at $3\times$ the compute.
Concretely, exploration improves FLOP efficiency by $4.1\times$, sample efficiency by $6.2\times$, parameter efficiency by $47\%$, lifts the strongest of image-generation recipes to a near-state-of-the-art 1.43 FID on ImageNet without guidance, enables \textit{scaling how end-to-end existing models are}, and unlocks \textit{scaling generalization}. Second, XMs enable end-to-end reconstructive generative modeling, matching diffusion on control tasks with $16$-$256\times$ fewer inference steps.
Together, these results establish XMs as both a \textit{new pretraining axis} for existing generative models and a standalone \textit{end-to-end generative modeling paradigm}.

\end{abstract}

%% file: main_sec/1_intro.tex
\epigraph{\textit{We scale the size of generative models and how much data we train them on... so why haven't we \textbf{scaled what they can generate?}}}{}

\section{Introduction}
\label{sec:intro}

AlexNet kicked off the deep learning revolution when \textit{end-to-end} training beat hand-designed layer-wise training, demonstrating that \textit{learning everything} performs better than hand-engineering~\cite{krizhevsky2012imagenet}. Since then, end-to-end neural networks have broadly replaced many hand-built pipelines across the field including image classification~\cite{he2016deep, krizhevsky2012imagenet}, object detection~\cite{carion2020end}, and image segmentation~\cite{kirillov2023segment}, learning each task directly from data.
Much of this success comes down to a single property: end-to-end models perform inference exactly as they were trained, so they are never exposed to inputs unlike those seen in training, which avoids distribution shifts and exposure bias that degrade performance and generalization~\cite{recht2019imagenet, hendrycks2019benchmarking, koh2021wilds, ranzato2016sequence, bengio2015scheduled}.

Despite this trend, generative modeling has remained the holdout: its most common and scalable recipes today---\textit{reconstructive} generative models (described in Section~\ref{sec:background})---are not end-to-end, sampling completely differently at inference than during training. For example, autoregressive and diffusion models are trained to predict a single step, but used at inference as recurrent neural networks over hundreds to thousands of predicted tokens or denoising steps, so per-step errors feed into the next step, drifting inputs off the training distribution and compounding errors~\cite{ranzato2016sequence, bengio2015scheduled, zhang2023snowball, ning2023elucidating, li2023alleviating}.
This shortcoming raises a simple question: \textit{``Why can't we train reconstructive generative models end-to-end?''} 

We argue this is because generative modeling is fundamentally about handling \textit{multimodal probability distributions}. To achieve this at scale, existing reconstructive models---whether autoregressive, diffusion~\cite{ho2020denoising}, or single-step~\cite{song2023consistency, geng2026mean}---factor the \textit{generation procedure} into smaller steps during training, making each step's target nearly unimodal so that a reconstruction loss no longer blurs distinct modes into their average. This factorization, however, is exactly what prevents current generative models from being end-to-end, so what else can we factor instead?

\begin{figure}[t]
    \centering
    \includegraphics[width=0.9\textwidth]{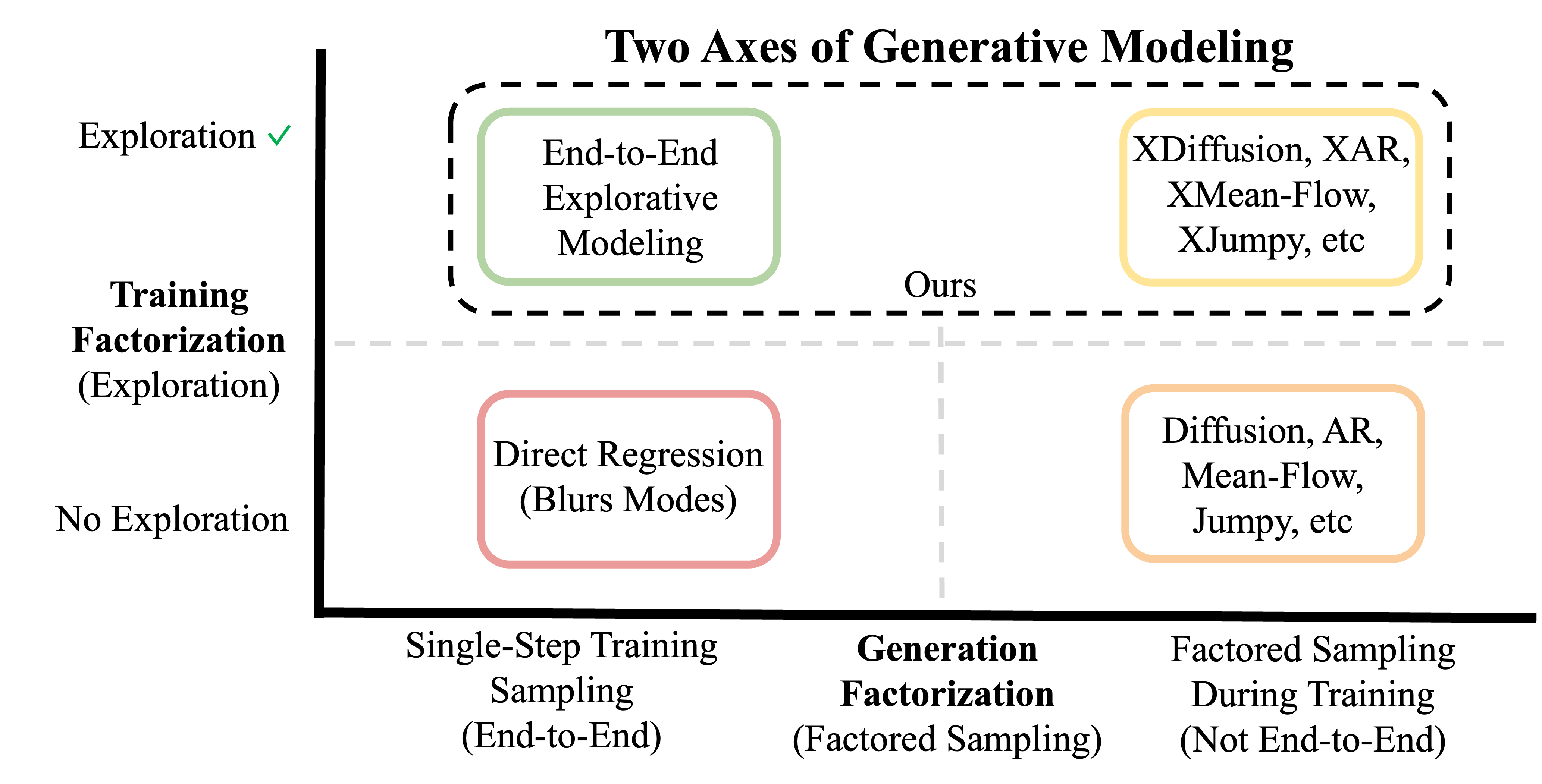}
    \caption{\textbf{Factorization Axes of Generative Modeling.} A generative model can factor either \textit{generation} (x-axis) or \textit{training} (y-axis). Factoring generation breaks sampling into many steps during training, making a model \textit{not} end-to-end (right column); factoring training involves \textit{exploration}, which trains on the modes a model captures best (top row). With neither, direct regression blurs distinct modes into their average (bottom left). Existing generative models factor generation but never training (bottom right), so adding exploration is a new pretraining axis for them (top right), while factoring training alone yields End-to-End Explorative Modeling (top left).}
    \label{fig:gen_modeling_axes}
    \vspace{-15pt}
\end{figure}

A generative model has only two processes to decompose---how it \textit{generates} and how it \textit{trains} (Figure~\ref{fig:gen_modeling_axes}). Because we've ruled out factoring generation, we factor the \textit{training loop} itself---a new paradigm we call \textbf{Explorative Modeling}: at each training step, the model explores $K$ possible matches between what it generates and the data, and trains on the closest. 
Because this happens entirely during training, Explorative Models (\textbf{XMs}) capture multimodal distributions while enabling end-to-end generation.

Exploration works by searching for which latent should be matched to which datapoint, something standard generation factorization completely sidesteps. In generative modeling, there is nothing that determines which datapoint each latent, such as input noise, should produce, so a latent is typically paired with targets at random. When models are trained to reconstruct many different valid targets at random, the best a single prediction can do is predict their average, a blur that matches no real datapoint (Figure~\ref{fig:exploration_improves_coverage} XM-1). 
XMs instead search for the latent whose generation is already closest to each datapoint, so each explored candidate can commit to a different mode, meaning the number of modes a model can capture, its \textit{generative expressivity} (Section~\ref{sec:background}), grows directly with the amount of exploration.

As a consequence, we find XMs are valuable in two settings. First, added to existing generative models, exploration is a \textit{new pretraining axis} beyond parameters and data. Existing generative models fix generative expressivity at training time through how they factor generation, so when they cannot capture every mode in the data, performance is capped no matter how far parameters and data scale. Because exploration scales generative expressivity directly, it relieves a bottleneck the other axes cannot, monotonically improving performance across both continuous and discrete domains---including images, video, and language.
Crucially, like scaling parameters or data,\footnote{Under compute-optimal scaling, parameters and data have to grow together: increasing one while holding the other fixed becomes increasingly suboptimal~\cite{hoffmann2022training}. Our findings show that exploration largely acts the same way: as scale increases, models without exploration fall increasingly short of compute-optimal performance.} these gains \textit{grow with scale} rather than saturate, rising from $7\%$ to $36\%$ as data grows, $13\%$ to $23\%$ as models grow, and with efficiency gains more than doubling at $3\times$ the compute, so these numbers likely understate the gains at larger scale. Concretely, exploration improves FLOP efficiency by $4.1\times$, sample efficiency by $6.2\times$, parameter efficiency by $47\%$, lifts the strongest of image-generation recipes to a near-state-of-the-art $1.43$ FID on ImageNet without guidance, enables a \textit{compute-generalization tradeoff} where more exploration improves generalization, and increasing exploration enables existing generative models to be trained more end-to-end.

Second, as a standalone approach, exploration enables \textit{end-to-end reconstructive generative modeling}. We find end-to-end XMs match Diffusion Policy~\cite{chi2023diffusion} on behavior cloning and Diffuser~\cite{janner2022planning} on goal-conditioned world modeling, while taking as little as a single forward pass in place of hundreds ($16$-$256\times$ fewer).

In summary, we make the following contributions:
\begin{itemize}
  \item We introduce \textbf{Explorative Modeling}, a new paradigm for handling multimodal distributions that works by factoring the \textit{training loop} instead of the generation procedure.
  \item We show exploration is a \textit{new scaling axis} for existing generative models, with gains in FLOP, parameter, and sample efficiency that \textit{increase with scale}.
  \item We find exploration enables a \textit{compute-generalization tradeoff}, where spending more training compute on exploration directly improves generalization.
  \item We show factorizing training and generation are \textit{substitutable}: as exploration increases, the optimal generative model becomes more end-to-end, indirectly improving generalization.
  \item We present a \textit{scalable end-to-end reconstructive generative modeling approach}, matching diffusion on control tasks at $16-256\times$ less inference compute.
\end{itemize}

Ultimately, exploration lets us scale how end-to-end existing generative models are, and taken to its limit, makes generative modeling fully end-to-end---extending to generative modeling the end-to-end training that has driven the rest of deep learning.

%% file: main_sec/2_background.tex
\section{Background}
\label{sec:background}

In standard supervised learning, such as classification or regression, each input generally has a single correct output, so a deterministic mapping is sufficient. Generative modeling has no such mapping: a request like ``generate a dog'' has no single right answer, as there are billions, or even infinitely many, valid dog images. These valid outputs are the \textit{modes} of the data distribution, and this large number of modes is what makes generation hard, so capturing them is the central focus of generative modeling.

\subsection{Mode Forcing}
\label{sec:mode_forcing}
\begin{figure}[t]
  \centering
  \captionsetup[subfigure]{font=scriptsize}

  \hspace*{0.205\columnwidth}%
  \begin{minipage}{0.785\columnwidth}
    \centering
    {\small\bfseries
    Increasing Exploration $K$ $\rightarrow$ Increases Generative Expressivity\par}
    \vspace{2pt}
    \begin{tikzpicture}[x=\linewidth,y=1pt]
      \draw[
        -{Latex[length=2mm,width=1.4mm]},
        line width=0.5pt
      ] (0,0) -- (1,0);
    \end{tikzpicture}
  \end{minipage}

  \vspace{4pt}

  \begin{subfigure}[b]{0.19\columnwidth}
    \includesvg[width=\linewidth]{fig/piles_3_ground_truth.svg}
    \caption{Ground Truth}
  \end{subfigure}\hfill
  \begin{subfigure}[b]{0.19\columnwidth}
    \includesvg[width=\linewidth]{fig/piles_3_xm_k1.svg}
    \caption{XM-1 (No Exploration)}
  \end{subfigure}\hfill
  \begin{subfigure}[b]{0.19\columnwidth}
    \includesvg[width=\linewidth]{fig/piles_3_xm_k2.svg}
    \caption{XM-2}
  \end{subfigure}\hfill
  \begin{subfigure}[b]{0.19\columnwidth}
    \includesvg[width=\linewidth]{fig/piles_3_xm_k5.svg}
    \caption{XM-5}
  \end{subfigure}\hfill
  \begin{subfigure}[b]{0.19\columnwidth}
    \includesvg[width=\linewidth]{fig/piles_3_xm_k50.svg}
    \caption{XM-50}
  \end{subfigure}

  \medskip

  \begin{subfigure}[b]{0.19\columnwidth}
    \includegraphics[width=\linewidth]{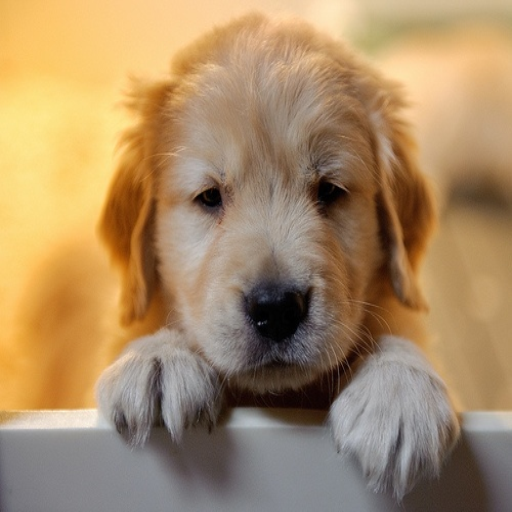}
    \caption{Ground Truth}
  \end{subfigure}\hfill
  \begin{subfigure}[b]{0.19\columnwidth}
    \includegraphics[width=\linewidth]{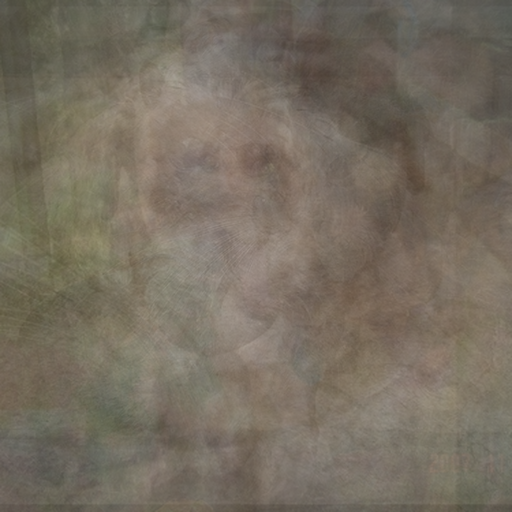}
    \caption{XM-1 (No Exploration)}
  \end{subfigure}\hfill
  \begin{subfigure}[b]{0.19\columnwidth}
    \includegraphics[width=\linewidth]{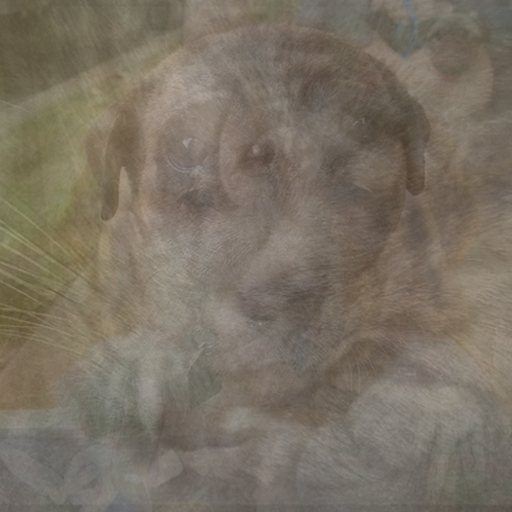}
    \caption{XM-5}
  \end{subfigure}\hfill
  \begin{subfigure}[b]{0.19\columnwidth}
    \includegraphics[width=\linewidth]{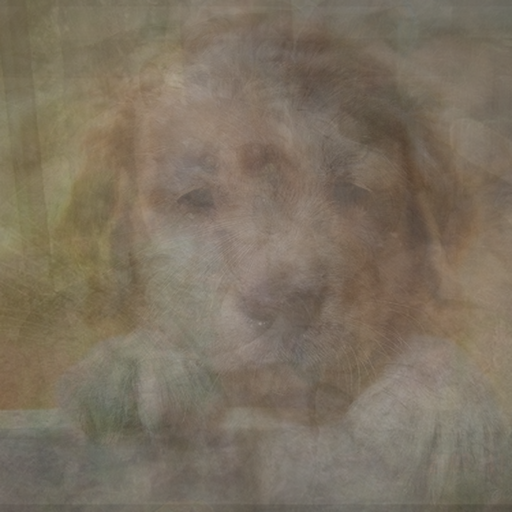}
    \caption{XM-20}
  \end{subfigure}\hfill
  \begin{subfigure}[b]{0.19\columnwidth}
    \includegraphics[width=\linewidth]{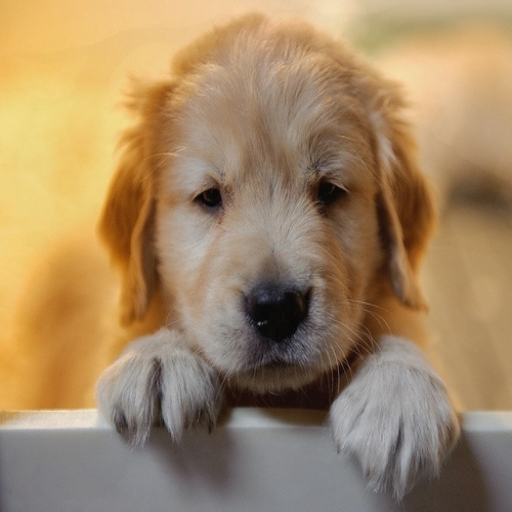}
    \caption{XM-50}
  \end{subfigure}

  \medskip

  \begin{subfigure}[b]{0.19\columnwidth}
    \includegraphics[width=\linewidth]{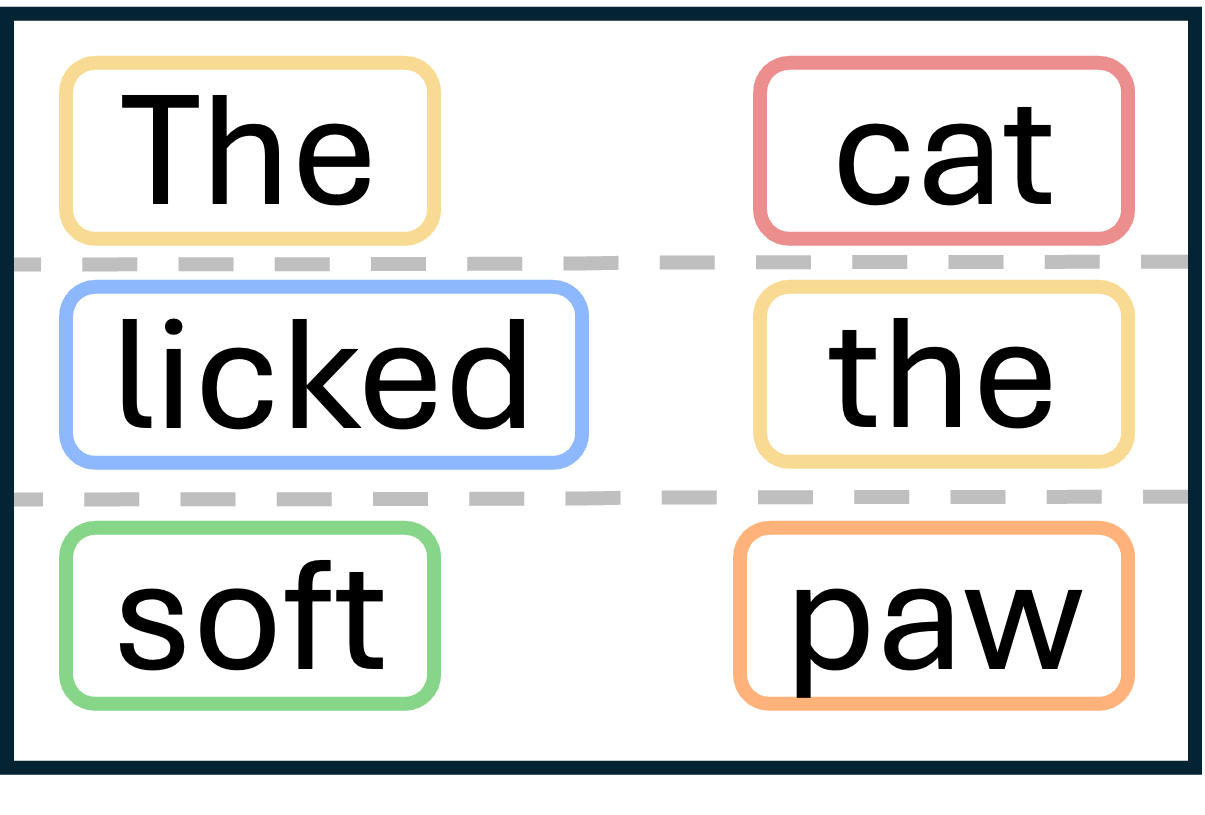}
    \caption{Ground Truth}
  \end{subfigure}\hfill
  \begin{subfigure}[b]{0.19\columnwidth}
    \includegraphics[width=\linewidth]{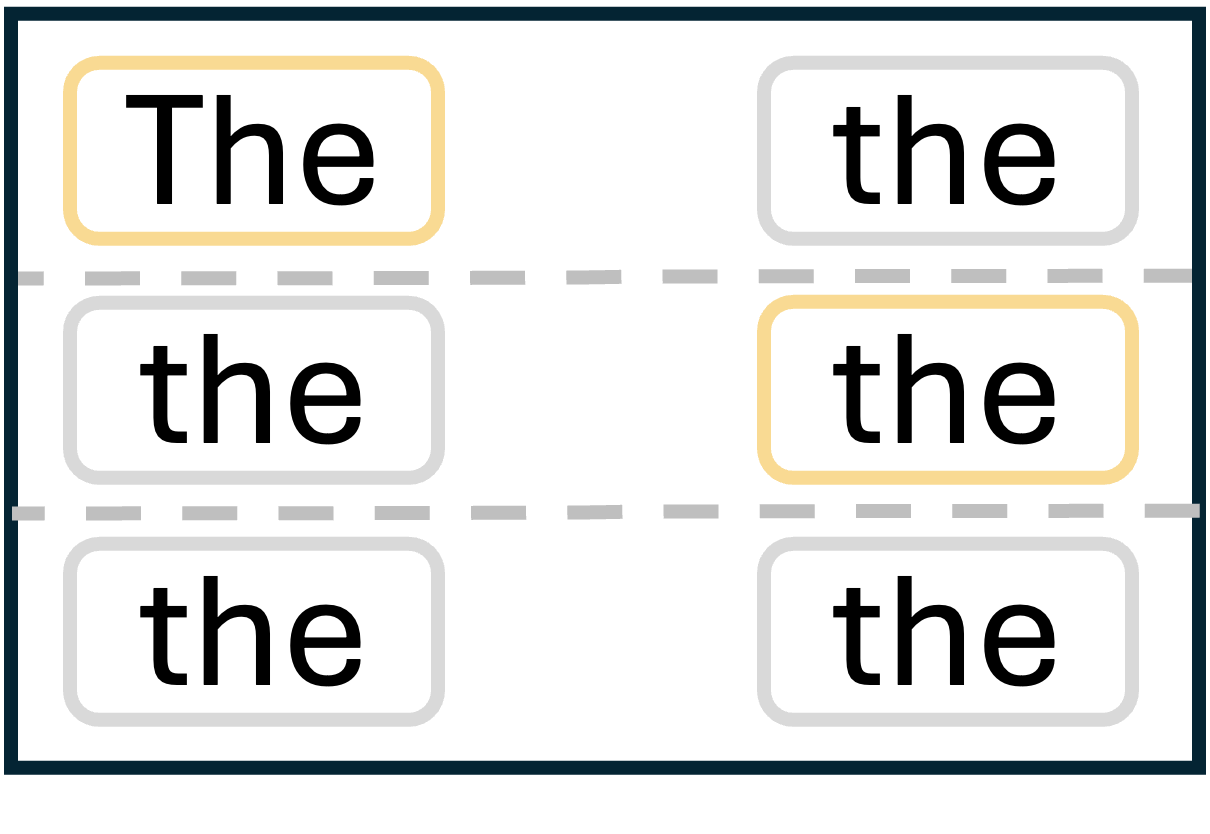}
    \caption{XM-1 (No Exploration)}
  \end{subfigure}\hfill
  \begin{subfigure}[b]{0.19\columnwidth}
    \includegraphics[width=\linewidth]{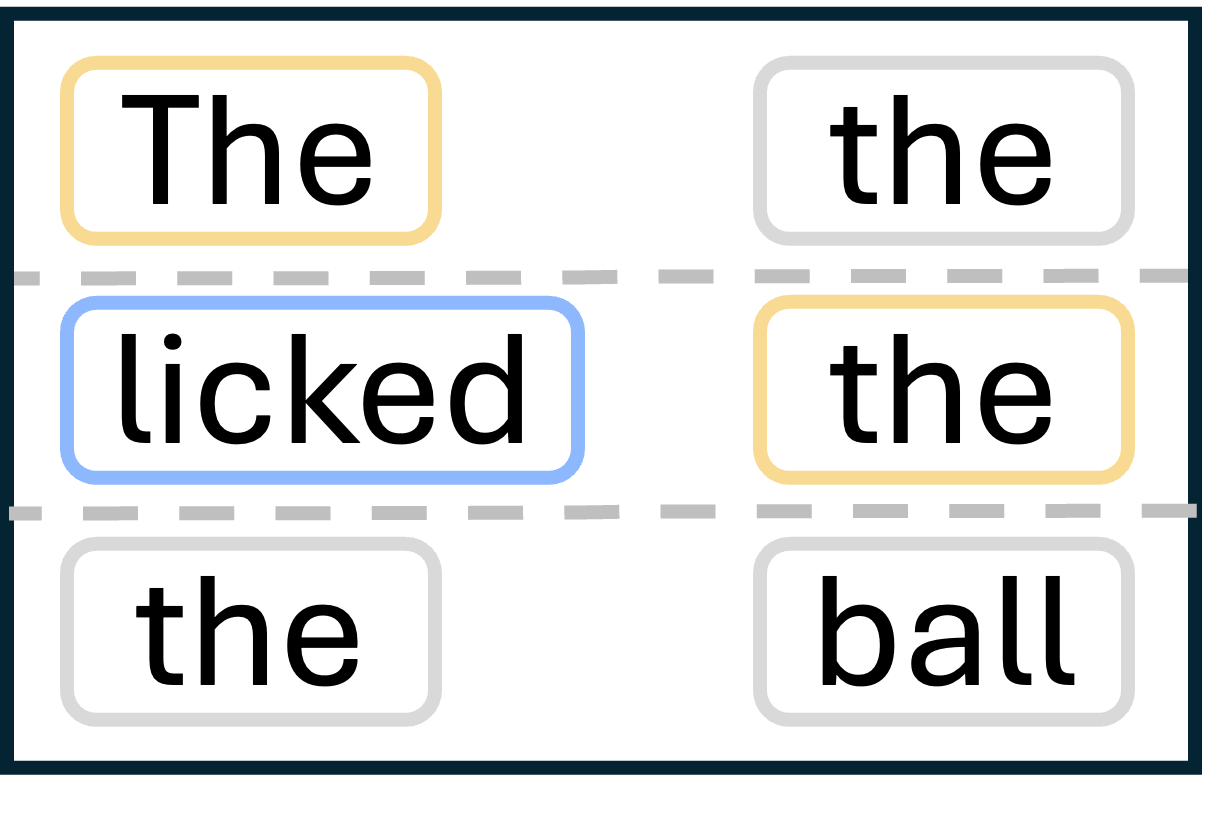}
    \caption{XM-2}
  \end{subfigure}\hfill
  \begin{subfigure}[b]{0.19\columnwidth}
    \includegraphics[width=\linewidth]{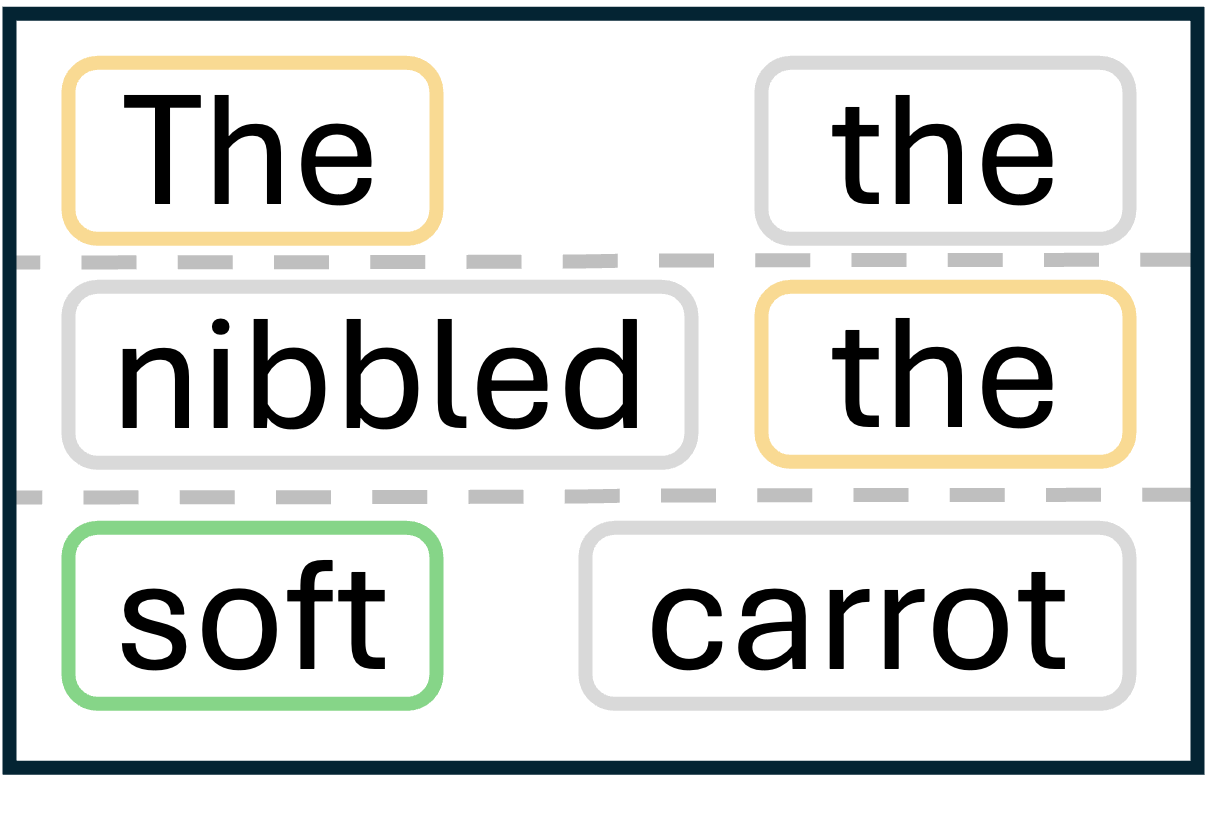}
    \caption{XM-4}
  \end{subfigure}\hfill
  \begin{subfigure}[b]{0.19\columnwidth}
    \includegraphics[width=\linewidth]{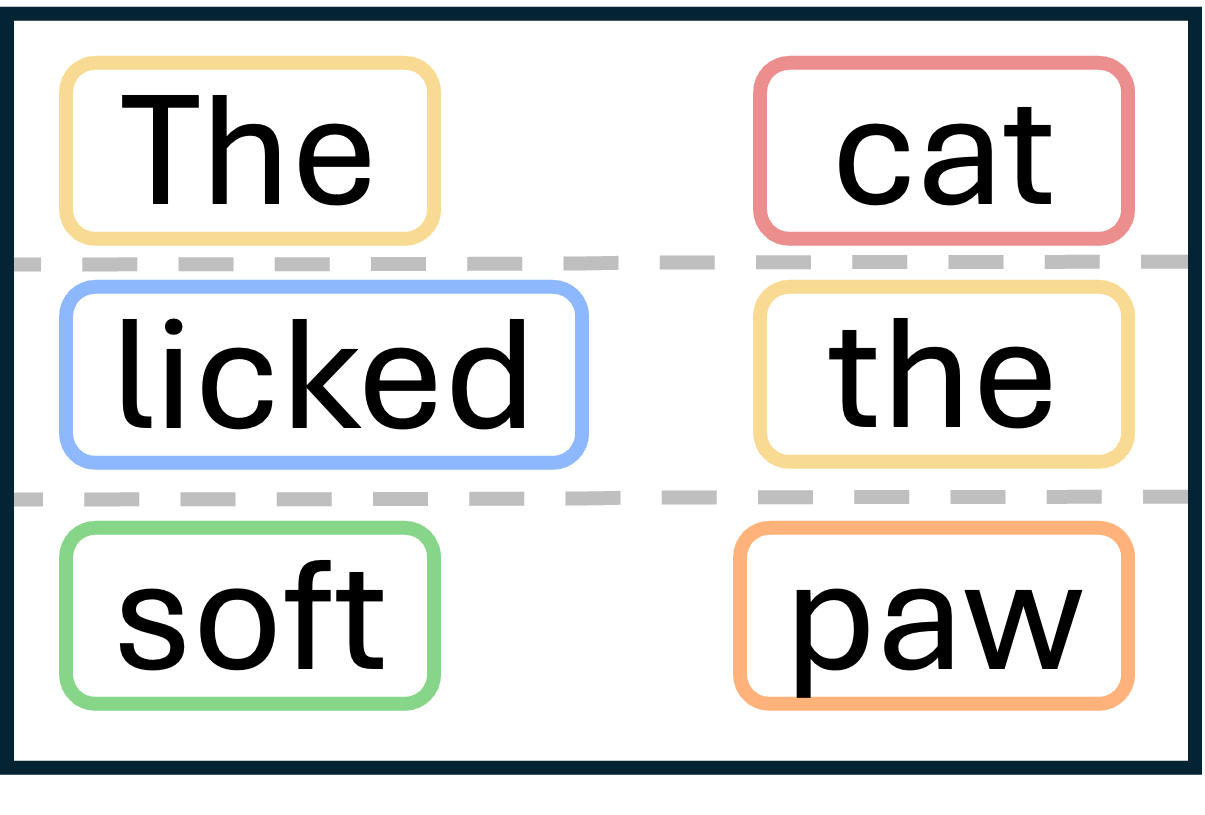}
    \caption{XM-8}
  \end{subfigure}

  \caption{
    \textbf{Increasing Exploration Scales Generative Expressivity and Reduces Blurring.}
    Each row shows trained model generations varying only the amount of
    exploration (XM-$K$, denoting XMs with $K$ modes explored):
    2D mixture generation (top), image generation (middle), and masked
    diffusion language modeling (MDLM)~\cite{sahoo2024simple} (bottom).
    With direct end-to-end regression (XM-1), models can only predict the
    mean of all samples---a single dot for three piles, a blurry image,
    and the word ``the'' repeated. XM-1 in the bottom row is the standard
    MDLM objective, which is prone to this collapse. As exploration
    increases, models become more \textit{generatively expressive},
    capturing the modes progressively better until generating
    high-quality samples in the right column.
  }
  \label{fig:exploration_improves_coverage}
  \vspace{-15pt}
\end{figure}

Generative models broadly fall into two families: \textit{reconstructive} and \textit{contrastive}~\cite{gladstone2026mode}. Contrastive generative models, such as GANs~\cite{goodfellow2014generative} and contrastive-divergence energy-based models (CD EBMs)~\cite{hinton2002training, du2019implicit}, are trained by contrasting generated data against true data---leveraging relative supervision from comparing samples with no explicit target---but have struggled with scalability. We therefore focus on \textit{reconstructive} generative models---the most common family that has scaled best thus far---which are trained by mapping a self-produced input, such as noise or a corrupted sample, back to an explicit data target that supervises each prediction, and include autoregressive, diffusion~\cite{ho2020denoising}, and flow~\cite{lipman2022flow} models. This pairing of an input with the target it should map to is called the \textit{coupling}, and the challenge with reconstructive models is that we do not know this coupling beforehand, so a single input is typically coupled to many valid targets across the dataset. This one-to-many coupling is what causes mode blurring when doing generative modeling naively, as the reconstruction loss minimizer of many targets is the mean, which lands between modes and matches no real datapoints (demonstrated in Figure~\ref{fig:exploration_improves_coverage} for XM-1).

Recent work on \textbf{Mode Forcing}~\cite{gladstone2026mode} points out that every scalable reconstructive model is built to dodge exactly this blur, with the central thesis that \textit{modern generative modeling is the art of designing a reconstructive objective whose loss minimizer captures modes instead of averaging them}. Existing approaches for achieving this at scale function by \textit{factoring generation} into a sequence of smaller, nearly unimodal steps, so no single prediction is forced to average across modes. For instance, autoregressive models reconstruct a target one element at a time, predicting each element from the ones already revealed, which means there is rich conditioning to make the prediction over the next element nearly unimodal. Diffusion and flow models~\cite{ho2020denoising, lipman2022flow} instead reveal the target gradually through denoising: each step conditions on a slightly noisier version and predicts a slightly cleaner one, keeping every step nearly unimodal. This \textit{factoring of the generation procedure} is why scalable generative modeling approaches succeed at generating high quality samples, whereas direct, single-step regression does not.

In general, scaling generative models has meant scaling just two axes: parameters or model size/FLOPs, which govern what a model can \textit{represent}, and the amount of data and training length, which govern what a model can \textit{learn}.
Mode Forcing suggests these two axes miss a third capacity:

\begin{definitionbox}
\textbf{Generative Expressivity:} The number of distinct modes a generative model's training objective allows it to capture.
    \end{definitionbox}

Unlike parameters and data, generative expressivity is set by the training objective itself, so it stays fixed no matter how far the other two axes scale. When an objective allows for capturing fewer modes than the data has, the surplus modes are not dropped but averaged, so a single prediction lands between them and matches no real datapoint (Figure~\ref{fig:exploration_improves_coverage} XM-1).
Formally, letting $M(q)$ denote the mode count of a distribution $q$ and $P_\theta(\cdot \mid c)$ the model's sampling distribution given conditioning $c$, generative expressivity is the largest conditional mode count an approach's loss minimizers can retain over any data distribution, $E \triangleq \sup_{p^*\!,\,c}\, \sup_{\theta^\star \in \arg\min_\theta \mathcal{L}(\theta)} M\big(P_{\theta^\star}(\cdot \mid c)\big)$ (further details in Section~\ref{sec:generative_expressivity_details}).
Direct regressors demonstrate why this axis matters, as they have $E=1$: even with \textbf{unlimited parameters and data}, their best possible output (loss minimizer) is still a single blurred mean of all the modes (Figure~\ref{fig:exploration_improves_coverage} XM-1).
This is counterintuitive because direct squared-error regression is itself maximum likelihood under a fixed-variance Gaussian---the likelihood is maximized faithfully, just over a density with a generative expressivity of one, whose best fit to multimodal data is the mean. Therefore, this suggests that \textit{the traditional notion of performing some form of maximum likelihood with sufficient data and parameters is not enough}, but rather that \textit{generative expressivity} is an overlooked scaling axis.
This may also explain why likelihood has long been observed to correlate poorly with sample quality~\cite{theis2016note}, as likelihood measures how well a density is fit while generative expressivity determines how many modes that density can hold. As a consequence, the field's primary goal of optimizing \textit{likelihood alone} may be the wrong one to chase, and generative expressivity should be optimized alongside it.
At its core, Explorative Modeling is a new way to increase generative expressivity (Figure~\ref{fig:gen_modeling_axes})---exploring $K$ candidates with a direct regressor raises generative expressivity to at least $K$, sharpening blurred means into distinct modes (Figure~\ref{fig:exploration_improves_coverage}).
Because factoring generation exists to supply this same quantity, factorizing generation and training are \textit{substitutable}, which we confirm empirically in Section~\ref{sec:experimentation}.

However, factoring generation does not remove this generative expressivity limitation entirely, as even highly scalable generative modeling approaches, such as diffusion and autoregression, can leave modes uncaptured when single predictions inside their factored procedure face many valid targets at once---that is, when the generative expressivity of those single predictions is too low.
Notably, this phenomenon \textit{worsens} with scale: as we add parameters and data, model expressivity and what models can learn cease to be the limiting factors, and generative expressivity increasingly becomes the bottleneck (we show this in Section~\ref{sec:experimentation}). Evidence for this already exists in how heavily today's models lean on guidance. Classifier-free guidance~\cite{ho2022classifier} sharpens samples by \textit{pushing them away} from the unconditional model, which, with less conditioning to pin down each prediction, blurs modes more than the conditional model. This extrapolation helps primarily because the original model itself blurs modes. Autoguidance~\cite{karras2024guiding} reinforces this perspective even further, improving samples by pushing away from a deliberately worse, more mode-collapsed version of the model, demonstrating that the \textit{core functionality of guidance is to push away from the conditional mean} due to challenges in capturing the true distribution. Together, this evidence points to the idea that even today's best generative models may improve from added generative expressivity, which we confirm in Section~\ref{sec:experimentation}.

\subsection{End-to-End Generation}
\label{sec:e2e_gen_modeling}

We call a generative model \textit{end-to-end} when it samples the same way during training and inference, so it is never exposed to inputs at inference that it was not trained on.\footnote{This concerns the model's own generation procedure, not test-time distribution shift---an end-to-end model can still face out-of-distribution test data, which is the ordinary generalization problem.} This is worth seeking for the same reason it has reshaped the rest of deep learning: ever since AlexNet~\cite{krizhevsky2012imagenet}, learning a task in an end-to-end manner has continued to beat splitting tasks into hand-designed stages. In generative modeling, this means learning the entire mapping from noise to data, including both its representations and its trajectory, rather than hand-specifying any part of it. End-to-end generative modeling also inherits the practical benefits of end-to-end training, where it removes \textit{exposure bias}~\cite{ranzato2016sequence, bengio2015scheduled, zhang2023snowball, ning2023elucidating, li2023alleviating} and the train-inference mismatch that drifts a model onto out-of-distribution inputs and compounds errors~\cite{recht2019imagenet, hendrycks2019benchmarking, koh2021wilds}, while as a byproduct enabling inference to be more efficient.

Yet general end-to-end \textit{reconstructive} generation remains unsolved.\footnote{Contrastive models such as GANs~\cite{goodfellow2014generative} and CD EBMs~\cite{hinton2002training,du2019implicit} have been end-to-end for a long time, but as noted in Section~\ref{sec:mode_forcing} they have struggled to scale.} Existing scalable reconstructive models capture modes by factoring generation into many steps, which makes training and inference inherently different---a diffusion model trained on a single denoising step is unrolled over hundreds of them at inference. Although recent methods push inference down to a single step~\cite{song2023consistency, geng2026mean}, during training they still anchor each prediction to the multi-step trajectory---conditioning on noised states or tying to the flow field---so that targets stay near-unimodal and avoid blurring. Their training therefore rarely simulates the one-step sampling they use at inference, so the train-inference mismatch remains and they are still not end-to-end.

\subsection{Existing Generative Models}
\label{sec:existing_gen_models}

To study Explorative Modeling as a new scaling axis for existing generative models, our experiments build on two main families of generative models. First, we build on top of \textbf{Diffusion and Flow Matching} models~\cite{ho2020denoising, lipman2022flow} which generate by repeatedly denoising via small steps from noise to data. 
Because Diffusion and Flow are equivalent formulations~\cite{gao2025diffusionmeetsflow}, we use the terms interchangeably throughout, and every Diffusion model over
continuous data in this work is trained with the Flow Matching objective, as Flow has generally performed best~\cite{ma2024sit}.
Second, we experiment with \textbf{Jumpy} generative models~\cite{gladstone2026jumpy}, which generalize the idea of Diffusion/Flow by varying the number of steps, or \textit{jumps}, interpolating between direct end-to-end regression (a single jump, the most end-to-end) and continuous-time flow (infinitely many jumps). This interpolation via the number of jumps enables a tradeoff between how end-to-end models are (fewer jumps) and how generatively expressive models are (more jumps). We return to this tradeoff when measuring how different models scale with exploration (Section~\ref{sec:experimentation}).

%% file: main_sec/3_approach.tex
\section{Explorative Modeling Approach}
\label{sec:approach}

\subsection{Explorative Modeling Intuition}
\label{sec:intuition}

The goal of all reconstructive generative models is to design a training objective such that the loss minimizer captures modes instead of averaging them. The reason for this goal is demonstrated in Figure~\ref{fig:exploration_improves_coverage} for XM-1, where performing end-to-end direct regression with a naive training objective generates samples that do not belong to the data distribution. Existing generative models achieve this goal by factoring the \textit{generation procedure} into a sequence of smaller steps, keeping each step's target nearly unimodal so that no single prediction is forced to average across modes (Section~\ref{sec:mode_forcing}). While these approaches have resulted in highly performant large-scale generative models~\cite{openai2023gpt4,rombach2022high}, they prevent models from being end-to-end (described in Section~\ref{sec:e2e_gen_modeling}), which directly hurts performance and generalization due to exposure bias~\cite{ranzato2016sequence, bengio2015scheduled, zhang2023snowball, ning2023elucidating, li2023alleviating}. Therefore, instead of factorizing the generation procedure, the goal of Explorative Modeling is to enable a \textit{new factorization axis}---the training loop itself (Figure~\ref{fig:gen_modeling_axes}).

In their simplest form, XMs are just \textit{best-of-$K$}, an idea that has appeared many times in prior work~\cite{lee2016stochastic, li2018implicit, vahabpour2024diverse} (Section~\ref{sec:related_work}). At each training step, the model generates $K$ candidate samples and trains on only the one closest to the data, implemented as a simple for loop (Algorithm~\ref{alg:min_samples}) where only the best generation receives gradients.
Formally, for a data sample $x$, generations $\hat{y}_1, \ldots, \hat{y}_K \sim G_\theta$, and a reconstruction loss $J$ such as squared error, the objective is
\begin{equation}
    \mathcal{L}(\theta) = \min_{i \in \{1, \ldots, K\}} J(\hat{y}_i, x).
    \label{eq:forward}
\end{equation}
Intuitively, the purpose of this for loop is to \textit{change the loss minimizer from the mean of the data samples toward the true data samples themselves}. This matters because for most data, the mean of samples is not on the data manifold, as demonstrated in Figure~\ref{fig:exploration_improves_coverage}. When the loss minimizer is the true data samples, models generate data that looks like real samples instead of a blurred, off-manifold average. Throughout this section we describe XMs as standalone models for intuition, though exploration can be added on top of existing generative models (Section~\ref{sec:xm_scaling_axis}).

There are several intuitions for what XMs are doing to enable handling multimodal distributions:

\vspace{-5pt}
\paragraph{Scalable Training of Latent Variable Models}
Explorative Modeling can be seen as conditioning the generator on a \textit{latent variable} (e.g., noise for diffusion models, or a learned embedding for language models) that resolves which of the many valid targets an input maps to: conditioned on the right latent, the one-to-many coupling (Section~\ref{sec:mode_forcing}) becomes one-to-one, so the target becomes unimodal, and the blur disappears. The challenge with latent variable models is that we do not know which latent goes with which datapoint in advance. Variational Autoencoders (VAEs)~\cite{kingma2013auto} learn an encoder to infer this latent, which adds a KL term and risks posterior collapse, and consequently VAEs have struggled to scale well as standalone generative models~\cite{vahdat2020nvae, xiao2021tackling}. Explorative Modeling instead recovers the pairing through \textit{exploration}, exploring possible matches between what it generates and the data and training on the closest. This trades extra training compute for end-to-end search of the latent variables.

\vspace{-5pt}
\paragraph{A Scalable Way to Resolve Coupling}
Reconstructive models must pair each input with a target, referred to as a \textit{coupling}, which we do not know in advance. Pairing at random is what ties one input to many targets, which blurs them (Section~\ref{sec:mode_forcing}). Computing a better coupling directly does not scale: exact optimal transport is cubic in the number of samples, and its minibatch approximations~\cite{tong2023improving} only match within a batch, a biased proxy for the true global pairing~\cite{fatras2020learning}. Instead of computing a global or minibatch coupling, Explorative Modeling searches for a coupling aligned with the model's own samples. Because each pairing keeps only the best match rather than forcing an assignment over a whole batch, and searches the model's own samples rather than fixed noise, the coupling avoids minibatch OT's bias and co-adapts with the model throughout training.

\begin{figure}[t]
    \centering
    \includegraphics[width=\textwidth]{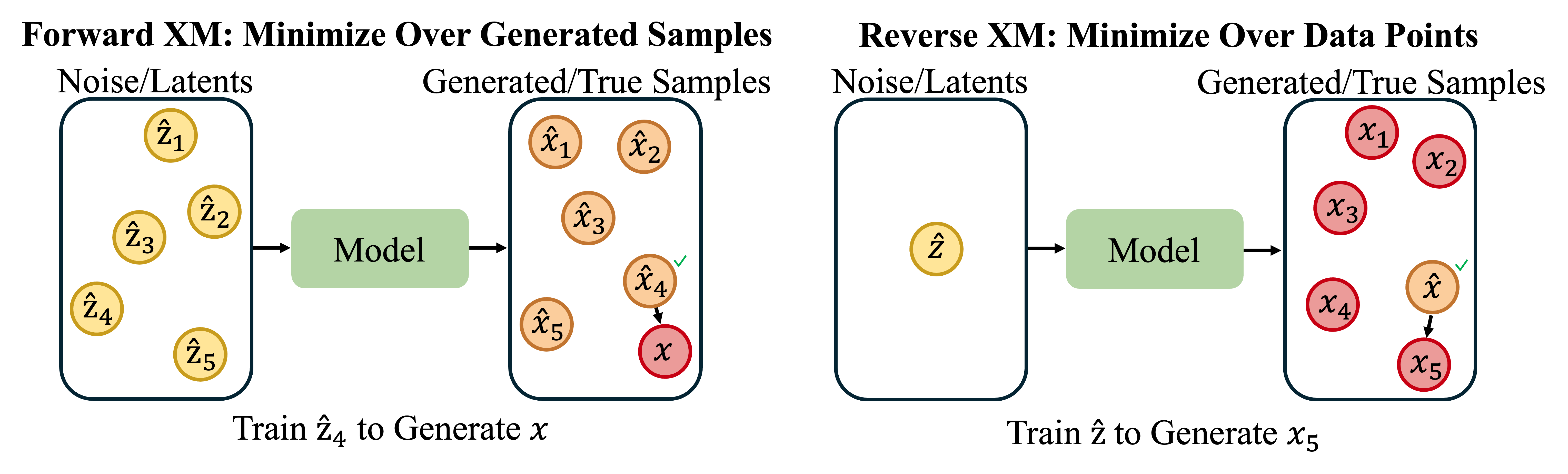}
    \caption{\textbf{Explorative Modeling Visualized.} Explorative Modeling explores possible matches between what the model generates and the data, and trains on the best match. This increases models' \textit{generative expressivity}, capturing multiple modes as opposed to predicting their mean (as in Figure~\ref{fig:exploration_improves_coverage} XM-1). In Forward XM, the model generates multiple samples that are compared to a ground truth sample. In Reverse XM, a generated sample is instead compared to many ground truth samples. In practice, both Forward and Reverse XM can be used together.}
    \label{fig:main_xm}
    \vspace{-15pt}
\end{figure}

\vspace{-5pt}
\paragraph{Generative Modeling via Search}
Explorative Modeling can be seen as recasting generative modeling as a \textit{search} problem, looking at training time for the latent, or coupling, that best explains the data. This framing is promising, as search and learning are the two methods the \textit{bitter lesson} identifies as scaling well with computation~\cite{sutton2019bitter}.

\vspace{-5pt}
\paragraph{Spreading Predictions Across Modes}
Geometrically, exploration changes what the best prediction strategy is. Consider guessing where darts land on a dartboard: with a single guess, the loss minimizer is the mean of all the throws, which is often a spot where few darts actually land. Forward XM (Equation~\ref{eq:forward}) instead makes $K$ guesses and scores only the closest, so the mean of the dartboard is no longer the loss minimizer---the best strategy becomes spreading the guesses so each covers a different cluster of throws. For the model, this means different latents specialize to different modes, so larger $K$ captures more modes instead of blurring them together (Figure~\ref{fig:exploration_improves_coverage}).

\vspace{-5pt}
\paragraph{Minimizing an Implicit Energy}
Exploration can also be seen as implicit training-time energy minimization, where the loss acts as an \textit{implicit energy} over pairings of a generation with the data, so searching for the lowest-loss match is a search for the minimum-energy, best-coupled samples. In this work, we search this landscape at random, but this search could instead be gradient-based (we describe this further in Section~\ref{sec:future_work}).

\subsection{Forward and Reverse Explorative Modeling}
\label{sec:forward_reverse_xm}

Exploration can search in either of two directions, which differ in what is held fixed and what is searched over (Figure~\ref{fig:main_xm}).

\vspace{-5pt}
\paragraph{Forward XM.} Forward XM fixes a data target and explores its own generations: it draws $K$ candidates and trains on the one closest to the target, exactly the best-of-$K$ objective of Equation~\ref{eq:forward} (Algorithm~\ref{alg:min_samples}), which we denote $\mathcal{L}_{\text{Forward}}$. 
Because every datapoint pulls in its nearest generation, no part of the data is ignored, so Forward XM is \textit{mass-covering}---it errs toward \textit{recall}, covering the full distribution. The challenge with Forward XM is compute---each of the $K$ candidates is a separate generation, so covering more modes takes more forward passes.

\vspace{-5pt}
\paragraph{Reverse XM.} Reverse XM fixes a generated model sample and searches the data: it draws a single sample $\hat y \sim G_\theta$ and trains it toward the closest of $K$ data targets $x_1, \dots, x_K \sim \mathcal{D}$ (Algorithm~\ref{alg:min_data}), flipping the objective to
\begin{equation}
\mathcal{L}_{\text{Reverse}}(\theta) = \min_{i \in \{1, \dots, K\}} J(\hat y, x_i).
\label{eq:reverse_xm}
\end{equation}
Each generation is pulled onto the data manifold, so Reverse XM errs toward \textit{precision}. Reverse XM is also cheap because it searches over data rather than generations, so each loss calculation only costs a single generation no matter how many targets it is compared against, which is useful for the large $K$ values needed to handle highly multimodal data. Reverse XM's weakness is that searching from the generation side applies no pressure to \textit{cover every mode}, so on its own it is \textit{mode-seeking} and can collapse onto a subset of the data. The two are therefore complementary---Forward XM focuses on recall/coverage whereas Reverse XM focuses on precision.
In practice, the two can be combined to control for precision and recall. Moreover, exploration can also be added onto existing generative models and applied to partial, masked, or noised samples rather than only full generations, as in the hybrid XMs of Section~\ref{sec:xm_scaling_axis}. We discuss implementation details for both variants in Appendix~\ref{sec:approach_details}.

\begin{figure}[t]
  \centering
  \begin{minipage}[t]{0.48\linewidth}
    \begin{algorithm}[H]
      \small
      \caption{Forward XM (Minimize over Generated Samples)}\label{alg:min_samples}
      \Inputs{Generator $G_\theta$, dataset $\mathcal{D}$, loss $J(\cdot)$}
      Sample $x \sim \mathcal{D}$\;
      \For{$i = 1, \dots, K$}{
        Sample $\hat y_i \sim G_\theta$\;
        $\mathcal{L}_i \gets J(\hat y_i, x)$\;
      }
      \Return{$\min_{i} \mathcal{L}_i$, update $\theta$}\;
    \end{algorithm}
  \end{minipage}\hfill
  \begin{minipage}[t]{0.48\linewidth}
    \begin{algorithm}[H]
      \small
      \caption{Reverse XM (Minimize over Data Points)}\label{alg:min_data}
      \Inputs{Generator $G_\theta$, dataset $\mathcal{D}$, loss $J(\cdot)$}
      Sample $\hat y \sim G_\theta$\;
      \For{$i = 1, \dots, K$}{
        Sample $x_i \sim \mathcal{D}$\;
        $\mathcal{L}_i \gets J(\hat y, x_i)$\;
      }
      \Return{$\min_{i} \mathcal{L}_i$, update $\theta$}\;
    \end{algorithm}
  \end{minipage}
  \vspace{-15pt}
\end{figure}

\vspace{-5pt}
\paragraph{What Forward and Reverse XM Optimize.} We can make the recall and precision behaviors of Forward and Reverse XM precise by asking what distribution each one drives the model toward. The starting point is that the squared error between a generation $\hat y$ and a datapoint $x$ is, up to a constant, the negative log of a Gaussian $k_\sigma(\hat y, x)$ of width $\sigma$ centered on the generation ($\sigma$ is an analysis device set by the loss scale, not a hyperparameter). Each generation can therefore be seen as placing a small bump of density around itself, and averaging these bumps over everything the model generates gives the model a density of its own,
\[
p_\theta(x) = \mathbb{E}_{\hat y \sim G_\theta}\!\left[k_\sigma(\hat y, x)\right] = (g_\theta * k_\sigma)(x),
\]
where $g_\theta$ is the distribution of the model's generations and $*$ is convolution, so $p_\theta$ is just $g_\theta$ blurred by the kernel. Blurring the data distribution $p^*$ the same way gives $p^*_\sigma = p^* * k_\sigma$. In their smooth, large-$K$ forms,\footnote{The smooth form scores the $K$ candidates by $-\log \tfrac{1}{K}\sum_i k_\sigma(\hat y_i, x)$ rather than by the best alone, and differs from the hard min by at most $\log K$.} Forward and Reverse XM then minimize
\[
\underbrace{\mathrm{KL}(p^* \,\|\, p_\theta) + H(p^*)}_{\text{Forward XM}}
\qquad\text{and}\qquad
\underbrace{\mathrm{KL}(g_\theta \,\|\, p^*_\sigma) + H(g_\theta)}_{\text{Reverse XM}},
\]
where $\mathrm{KL}$ measures the mismatch between two distributions and $H$ is entropy. The two objectives are mirror images except for the entropy each carries. Forward's entropy is the \textit{data}'s, a constant the model cannot change, so \textbf{Forward XM is maximum likelihood} of the mixture its explored candidates form, and its large-$K$ optimum recovers the data distribution up to that blur. Notably, this maximum likelihood reading holds at \textit{every} $K$; what changes with $K$ is the density the likelihood is fit over. At $K{=}1$ that density is a single Gaussian---the familiar fact that squared-error regression is Gaussian maximum likelihood---so the best the model can do is fit the blurred mean (Figure~\ref{fig:exploration_improves_coverage}), while at larger $K$ it is a mixture of the $K$ candidates that can hold $K$ modes, one per candidate. In other words, $K$ scales how many modes the density can capture (its generative expressivity), which is why maximum likelihood alone can be misleading (Section~\ref{sec:mode_forcing}). \textbf{Reverse XM targets the reverse KL}, the mode-seeking direction. However, the entropy in its objective is the \textit{model}'s own, so the model can lower its loss simply by shrinking its spread, potentially resulting in collapse. One solution is to add an entropy bonus that cancels the model's entropy term and leaves the pure reverse KL; another is to combine Reverse XM with Forward XM, which is mass-covering. We give precise statements, assumptions, and proof sketches for these claims in Appendix~\ref{sec:additional_theory}.

%% file: main_sec/4_experimentation.tex
\section{Experimentation and Results}
\label{sec:experimentation}

The goal of this section is to demonstrate that Explorative Modeling can be used both as a new pretraining axis (Section~\ref{sec:xm_scaling_axis}) as well as a standalone generative modeling approach (Section~\ref{sec:xm_e2e}). As a new pretraining axis, we experiment with generative modeling hybrids combining exploration with either Diffusion/Flow~\cite{sohl2015deep,ho2020denoising,lipman2022flow} or Jumpy~\cite{gladstone2026jumpy} generative models across both continuous and discrete domains. 
We refer to a generative model paired with Explorative Modeling by prefixing its name with an \textbf{X} (e.g., XDiffusion, XJumpy), reflecting how these are hybrid explorative and existing generative modeling combinations. Across all hybrid experiments, we do no XM-specific hyperparameter tuning, keeping each baseline recipe's hyperparameters unchanged and only adding exploration. As a standalone generative modeling approach, we compare XMs to strong baselines in Behavior Cloning and Goal-Conditioned World Modeling. For all experiments in the main section of this paper, we denote exploring $K$ modes as XM-$K$, and we use Forward XM (Section~\ref{sec:approach}), as it is simpler to implement (this is discussed further in Section~\ref{sec:future_work}). Note that all baselines without exploration are equivalent to XM-1, as exploring a single mode reduces to standard training. By default, all experiments in this section are without guidance, except for guidance-based results reported in Table~\ref{tab:rae_comparison}. Our largest image generation experiments report $\mathrm{FDr}^{6}$ because FID has saturated at this performance level~\cite{yang2026representation, singh2026improved}; $\mathrm{FDr}^{6}$ works by averaging the Fr\'echet distance to the training data over six representation spaces.

\subsection{Explorative Modeling as a New Scaling Axis}
\label{sec:xm_scaling_axis}

\begin{figure}[t]
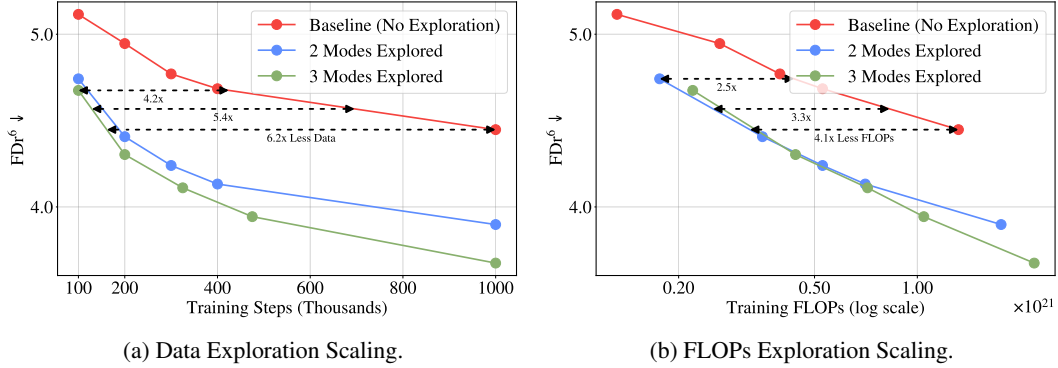

  \centering
  \begin{subfigure}[b]{0.49\columnwidth}
    \includesvg[width=\linewidth]{fig/xm_rae_fdr_scaling_ema_longer_comparisons_3_modes.svg}
    \caption{Data Exploration Scaling.}
    \label{subfig:xrae_data_eff}
  \end{subfigure}\hfill
  \begin{subfigure}[b]{0.49\columnwidth}
    \includesvg[width=\linewidth]{fig/xm_rae_fdr_scaling_ema_flops_longer_comparisons_3_modes.svg}
    \caption{FLOPs Exploration Scaling.}
    \label{subfig:xrae_flop_eff}
  \end{subfigure}
  \caption{\textbf{Exploration Improves Data and FLOP Efficiency at Scale.} We add exploration to RAE~\cite{zheng2025representation}, the state-of-the-art image generation recipe as of three months before this work (we report $\mathrm{FDr}^{6}$, as FID at this performance level is saturated~\cite{yang2026representation, singh2026improved}). Exploration reaches the baseline's best performance with $6.2\times$ less data (Figure~\ref{subfig:xrae_data_eff}) and $4.1\times$ fewer FLOPs (Figure~\ref{subfig:xrae_flop_eff})---more than doubling the gains of similar experiments using a third of the compute (Figure~\ref{fig:xm_flop_data_scaling})---demonstrating that gains from exploration grow with scale.}
  \label{fig:xm_rae_flop_step_eff}
  \vspace{-5pt}
\end{figure}

\begin{figure}[t]
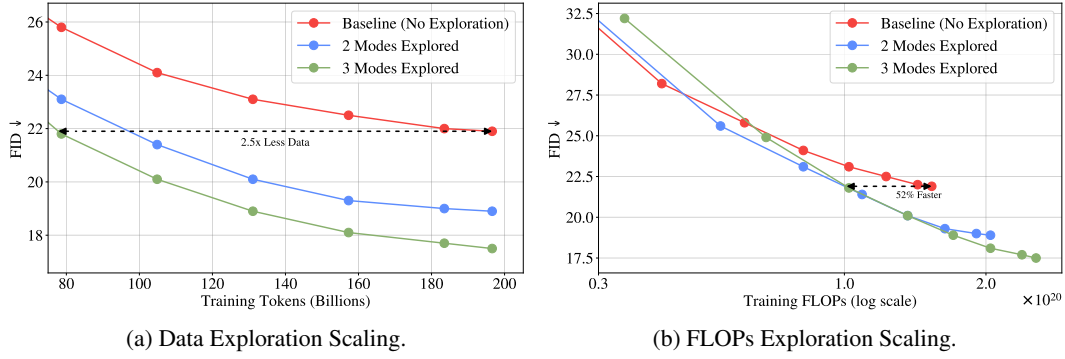

  \centering
  \begin{subfigure}[b]{0.49\columnwidth}
    \includesvg[width=\linewidth]{fig/xm_fid_steps_long_data.svg}
    \caption{Data Exploration Scaling.}
    \label{subfig:fid_xm_data_scaling}
  \end{subfigure}\hfill
  \begin{subfigure}[b]{0.49\columnwidth}
    \includesvg[width=\linewidth]{fig/xm_fid_flops_long.svg}
    \caption{FLOPs Exploration Scaling.}
    \label{subfig:fid_xm_flops_scaling}
  \end{subfigure}
  \caption{\textbf{Exploration Improves Sample and FLOP Efficiency.} We add exploration to an optimally tuned SiT baseline~\cite{ma2024sit}, training at roughly a third of the compute of Figure~\ref{fig:xm_rae_flop_step_eff}. Exploration reaches the same performance with $2.5\times$ less data (Figure~\ref{subfig:fid_xm_data_scaling}) and improves FLOP efficiency by as much as $52\%$ (Figure~\ref{subfig:fid_xm_flops_scaling}), with the compute-optimal amount of exploration increasing as models train longer---a trend that also holds at larger scale (Figure~\ref{subfig:xrae_flop_eff}).}
  \label{fig:xm_flop_data_scaling}
  \vspace{-20pt}
\end{figure}

\paragraph{Does Exploration Improve Existing Generative Models' Performance?}
Progress in generative modeling has largely been driven by scaling \textit{parameter expressivity} through training larger models. This raises the question---if scaling parameter expressivity helps, \textit{why not also scale \textbf{generative expressivity}, a model's capacity to capture multiple modes, rather than average them?} Existing scalable reconstructive generative models rely on the same generation factorization regardless of model size, fixing generative expressivity at design time rather than scaling it. If this factorization is not sufficient to capture all the modes in the distribution, we could expect that adding exploration, as a way to increase generative expressivity, could improve performance.
To investigate this, we experiment with adding exploration to existing generative models, including Diffusion/Flow and Jumpy generative models~\cite{gladstone2026jumpy} (which generalize Diffusion/Flow by varying the number of steps, or \textit{jumps}, interpolating between single-step regression and continuous-time Flow; Section~\ref{sec:existing_gen_models}).

\begin{wrapfigure}{r}{0.45\textwidth}
    \centering
    \includesvg[width=\linewidth]{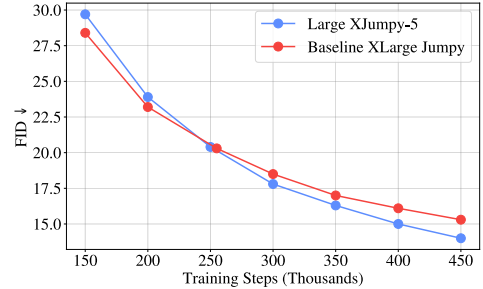}
    \caption{\textbf{Exploration Improves Parameter Efficiency.} A Large model with 5 modes explored \textit{scales better} than an XLarge model with 47\% more parameters and no exploration, demonstrating how exploration can improve parameter efficiency.}
    \label{fig:xm_param_eff}
    \vspace{-15pt}
\end{wrapfigure}

\begin{figure}[t]
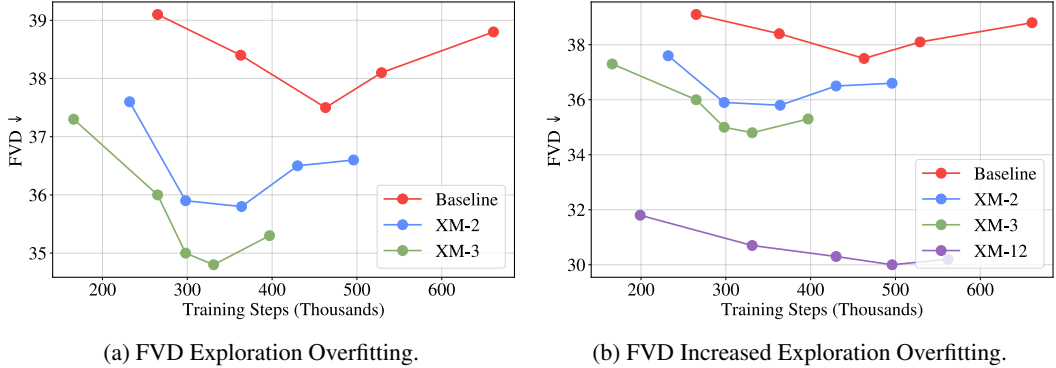

  \centering
  \begin{subfigure}[b]{0.49\columnwidth}
    \includesvg[width=\linewidth]{fig/xm_fvd_data_eff_jumpy.svg}
    \caption{FVD Exploration Overfitting.}
    \label{subfig:fvd_xm_explore_overfit}
  \end{subfigure}\hfill
  \begin{subfigure}[b]{0.49\columnwidth}
    \includesvg[width=\linewidth]{fig/xm_fvd_data_eff_jumpy_all.svg}
    \caption{FVD Increased Exploration Overfitting.}
    \label{subfig:fvd_xm_explore_overfit_more}
  \end{subfigure}
  \caption{\textbf{Exploration Improves Generalization.} As the number of modes explored increases, 4-step XJumpy models achieve a better absolute minimum FVD due to overfitting less. The right panel shows the same runs as the left, adding the most-explored model (XM-12) to show this trend continues to the maximum exploration we test. Models overfit in this setting due to training on the relatively small Something-Something V2 dataset~\cite{goyal2017something} (more on the setup in Section~\ref{sec:experimental_details}).
  }
  \label{fig:fvd_xm_overfitting}
  \vspace{-10pt}
\end{figure}

We begin by adding exploration to a strong image generation recipe (RAE~\cite{zheng2025representation}), training models that differ only in the amount of exploration. We find that exploration significantly improves performance throughout training, reaching the no-exploration baseline's final performance with $6.2\times$ less data\footnote{Throughout this paper, we refer to two notions of sample/data efficiency. Here we mean the first, or the number of training samples processed (training steps at a fixed batch size) to reach a given performance. The second notion is the best performance achievable on a fixed-size dataset before overfitting, which concerns generalization; we test that separately in Figure~\ref{fig:fvd_xm_overfitting}.} and $4.1\times$ fewer FLOPs (Figure~\ref{fig:xm_rae_flop_step_eff}).
These gains also hold beyond a single recipe, where adding exploration to an optimally tuned SiT baseline~\cite{ma2024sit} improves FLOP efficiency by as much as $52\%$ and reaches the same performance with $2.5\times$ less data (Figure~\ref{fig:xm_flop_data_scaling}). Notably, the SiT experiments use roughly a third of the compute of the RAE experiments, meaning the efficiency gains from exploration \textit{more than doubled} when moving to the larger-scale setting at $3\times$ the compute---suggesting gains from exploration grow with scale, a pattern we examine more directly below.

Notably, each explored mode in Forward XM adds compute (Reverse XM largely avoids this), yet despite this cost, the FLOP-optimal number of modes to explore grows as training continues (Figures~\ref{subfig:xrae_flop_eff} and~\ref{subfig:fid_xm_flops_scaling}), as generative expressivity increasingly becomes the bottleneck. This mirrors compute-optimal parameter scaling~\cite{hoffmann2022training}, where just as the optimal number of parameters grows as compute increases, the optimal amount of exploration grows too, meaning \textit{models that explore more modes eventually scale faster}. Exploration also improves parameter efficiency---a Large model exploring 5 modes outscales an XLarge model with $47\%$ more parameters and no exploration (Figure~\ref{fig:xm_param_eff}). These results show exploration is not just more performant, but that it enables a more efficient use of compute, data, and parameters.

\begin{takeaway}
\textbf{Takeaway:} Exploration improves existing generative models' FLOP efficiency by $4.1\times$, sample efficiency by $6.2\times$, and parameter efficiency by $47\%$, with efficiency gains more than doubling when compute is tripled.
\end{takeaway}

\begin{figure}[t]
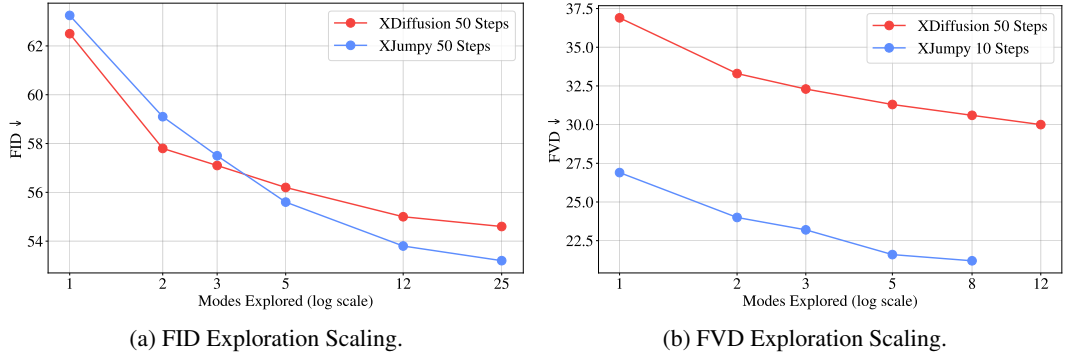

  \centering
  \begin{subfigure}[b]{0.49\columnwidth}
    \includesvg[width=\linewidth]{fig/xm_fid_overlap.svg}
    \caption{FID Exploration Scaling.}
    \label{subfig:fid_xm_explore_scaling}
  \end{subfigure}\hfill
  \begin{subfigure}[b]{0.49\columnwidth}
    \includesvg[width=\linewidth]{fig/xm_fvd_overlap.svg}
    \caption{FVD Exploration Scaling.}
    \label{subfig:fvd_xm_explore_scaling}
  \end{subfigure}
  \caption{\textbf{Increasing Exploration Monotonically Improves Performance.} As the number of modes explored increases, both FID (left) and FVD (right) improve monotonically for Explorative Diffusion (XDiffusion) and Explorative Jumpy (XJumpy) models. In both cases, XJumpy benefits more from exploration than XDiffusion, a gap we examine in more detail below.}
  \label{fig:xm_explore_fid_fvd_scaling}
  \vspace{-5pt}
\end{figure}

\begin{figure}[t]
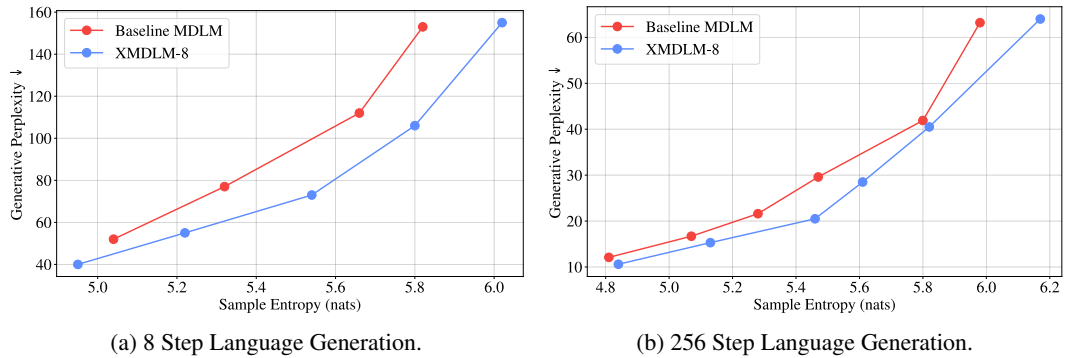

  \centering
  \begin{subfigure}[b]{0.49\columnwidth}
    \includesvg[width=\linewidth]{fig/xmdlm_8steps_v2.svg}
    \caption{8 Step Language Generation.}
    \label{subfig:mdlm_xm_scaling_8}
  \end{subfigure}\hfill
  \begin{subfigure}[b]{0.49\columnwidth}
    \includesvg[width=\linewidth]{fig/xmdlm_256steps_v2.svg}
    \caption{256 Step Language Generation.}
    \label{subfig:mdlm_xm_scaling_256}
  \end{subfigure}
  \caption{\textbf{Exploration Improves Masked Diffusion Language Modeling Performance.} Switching from a baseline Masked Diffusion Language Model (MDLM)~\cite{sahoo2024simple, lou2023discrete} to an Explorative MDLM (XMDLM) by exploring $8$ modes significantly improves performance, achieving a better Perplexity-Entropy frontier for all points. This demonstrates exploration can improve generative models in both discrete and continuous spaces.}
  \label{fig:mdlm_xm_scaling}
  \vspace{-15pt}
\end{figure}

\paragraph{Does Performance Across Modalities Scale with Exploration?} 
Having seen exploration added to existing models improve image generation performance, we next ask whether these benefits extend across modalities, and how they scale with the amount of exploration. To test this, we train image generation models, video generation models, and language models with a fixed parameter count, varying only the number of modes explored.
For both image and video generation, increasing exploration monotonically improves performance as measured by FID and FVD respectively (Figure~\ref{fig:xm_explore_fid_fvd_scaling}), with some models seeing a greater than 20\% performance boost.
This benefit carries over to discrete data, where adding exploration to a masked diffusion language model (MDLM) improves its perplexity-entropy frontier\footnote{This frontier has become the standard evaluation in this setting, as generative perplexity alone can be gamed by low-entropy sampling~\cite{zheng2024masked, pynadath2026generative}.} across the board (Figure~\ref{fig:mdlm_xm_scaling}), demonstrating that exploration helps in both continuous and discrete spaces. Notably, these gains \textit{do not stop} as exploration increases (Figure~\ref{fig:xm_explore_fid_fvd_scaling}), suggesting increased exploration could further improve performance.

\begin{takeaway}
\textbf{Takeaway:} Increasing exploration monotonically improves performance in both continuous and discrete spaces, for image, video, and language generation.
\end{takeaway}

\paragraph{Does Exploration Improve Generalization?}
So far, we have measured sample/data efficiency as the number of training samples a model must process to reach a target performance.
A stronger, more generalization-focused evaluation asks how much a model can extract from a \textit{fixed} dataset---its best achievable performance before it begins to overfit~\cite{prabhudesai2026diffusion}. We experiment with this setup, training models until their validation-set performance starts to get worse.
One reason a model may overfit comes down to its generative expressivity---its capacity to represent multiple modes rather than collapse them to an average (Section~\ref{sec:mode_forcing}). With limited generative expressivity, a model's best possible prediction is a blurred compromise between modes, which typically lies off the data manifold and matches no real datapoint, so a model fitting this compromise is memorizing something that does not exist in the true data distribution rather than generalizing. Even when generative expressivity is not the bottleneck, having surplus expressivity may ease optimization toward simpler solutions explaining the data, which tend to generalize better, much as overparametrization does~\cite{nakkiran2021deep,wilson2025deep}. Therefore, as exploration increases generative expressivity directly, it may improve generalization. 
We find this for video generation (Figure~\ref{fig:fvd_xm_overfitting}), where increasing exploration improves generalization by reducing overfitting, resulting in a better absolute minimum FVD ($30.0$ with exploration versus $37.5$ without exploration). Because this improvement comes from spending more training compute on exploration, it amounts to a \textit{compute-generalization tradeoff}: extra compute directly buys better generalization.
As data, rather than compute, increasingly becomes the bottleneck for large-scale training~\cite{prabhudesai2026diffusion, slowrun_2026, kim2025pretraining}, we see improved generalization as an especially promising characteristic of XMs.

\begin{takeaway} 
\textbf{Takeaway:} More exploration reduces overfitting, reaching better performance on a fixed dataset. This enables a \textit{compute-generalization tradeoff}, where extra training compute directly buys generalization.
\end{takeaway}

\paragraph{Does Exploration Improve State-of-the-Art Recipes at Scale?} If exploration is a genuine scaling axis, it should improve even the strongest, most heavily tuned recipes. 
The Representation Autoencoder (RAE)~\cite{zheng2025representation} recipe from Figure~\ref{fig:xm_rae_flop_step_eff} provides such a test: RAE was the state-of-the-art ImageNet 256$\times$256 image generation recipe as of three months before the release of this work, primarily involving a change of representation space from the SD-VAE~\cite{stabilityai_sd_vae, rombach2022high} to a Representation Autoencoder. Aside from the previously discussed $6.2\times$ data and $4.1\times$ FLOP efficiency improvements, we find the performance gains also hold as models converge (training XL models for up to $2.2\times10^{21}$ FLOPs), where an XM-2 RAE model reaches near-state-of-the-art non-CFG FID without post-training (Table~\ref{tab:rae_comparison}), and much better $\mathrm{FDr}^{6}$ than the baseline. Convergence also compounds across recipes: XRAE converges $6.2\times$ faster than RAE, which itself converges $47\times$ faster than SiT~\cite{zheng2025representation, ma2024sit}---making XRAE almost $300\times$ faster to converge than the standard SiT recipe.

\begin{table}[t]
\centering
\small
\renewcommand{\arraystretch}{1.1}
\setlength{\tabcolsep}{5pt}
\resizebox{\columnwidth}{!}{
\begin{tabular}{l ccccc ccccc}
\toprule
\multirow{2}{*}{\textbf{Method}} & \multicolumn{5}{c}{\textbf{Generation@256 w/o guidance}} & \multicolumn{5}{c}{\textbf{Generation@256 w/ guidance}} \\
\cmidrule(lr){2-6} \cmidrule(lr){7-11}
 & \textbf{FDr$^{6}$}$\downarrow$ & \textbf{gFID}$\downarrow$ & \textbf{IS}$\uparrow$ & \textbf{Prec.}$\uparrow$ & \textbf{Rec.}$\uparrow$ & \textbf{FDr$^{6}$}$\downarrow$ & \textbf{gFID}$\downarrow$ & \textbf{IS}$\uparrow$ & \textbf{Prec.}$\uparrow$ & \textbf{Rec.}$\uparrow$ \\
\midrule
\multicolumn{11}{l}{\textit{Latent Diffusion with VAE}} \\
\arrayrulecolor{black!30}\midrule
DiT~\cite{peebles2023scalable} & - & 9.62 & 121.5 & 0.67 & 0.67 & - & 2.27 & 278.2 & \textbf{0.83} & 0.57 \\
MaskDiT~\cite{zheng2023maskdit} & - & 5.69 & 177.9 & 0.74 & 0.60 & - & 2.28 & 276.6 & 0.80 & 0.61 \\
SiT~\cite{ma2024sit} & - & 8.61 & 131.7 & 0.68 & 0.67 & - & 2.06 & 270.3 & 0.82 & 0.59 \\
MDTv2~\cite{gao2023mdtv2} & - & - & - & - & - & - & 1.58 & \textbf{314.7} & 0.79 & 0.65 \\
VA-VAE~\cite{yao2025vavae} & - & 2.17 & 205.6 & 0.77 & 0.65 & - & 1.35 & 295.3 & 0.79 & 0.65 \\
REPA~\cite{yu2025repa} & - & 5.78 & 158.3 & 0.70 & 0.68 & - & 1.29 & 306.3 & 0.79 & 0.64 \\
DDT~\cite{wang2025ddt} & - & 6.27 & 154.7 & 0.68 & \textbf{0.69} & - & 1.26 & 310.6 & 0.79 & 0.65 \\
REPA-E~\cite{leng2025repae} & - & 1.70 & 217.3 & 0.77 & 0.66 & - & \textbf{1.15} & 304.0 & 0.79 & 0.66 \\
\arrayrulecolor{black}\midrule
\multicolumn{11}{l}{\textit{Latent Diffusion with RAE~\cite{zheng2025representation}}} \\
\arrayrulecolor{black!30}\midrule
$\text{DiT}^{\text{DH}}$-XL (DINOv2-B~\cite{oquab2023dinov2}) & 4.42 & 1.55 & 237.3 & \textbf{0.79} & 0.64 & 3.33 & 1.16 & 257.8 & 0.78 & \textbf{0.67} \\
\arrayrulecolor{black}\midrule
\multicolumn{11}{l}{\textit{RAE Recipe with Exploration (Ours)}} \\
\arrayrulecolor{black!30}\midrule
$\text{XDiT}^{\text{DH}}$-XL (DINOv2-B~\cite{oquab2023dinov2}), XM-2 & \textbf{3.91} & \textbf{1.43} & \textbf{240.3} & \textbf{0.79} & 0.64 & \textbf{3.17} & 1.19 & 254.9 & 0.77 & \textbf{0.67} \\
\arrayrulecolor{black}\bottomrule
\end{tabular}
}
\caption{\textbf{Exploration Improves State-of-the-Art Image Generation Recipes.} We add exploration to the Representation Autoencoder (RAE) recipe~\cite{zheng2025representation} used for ImageNet 256x256. On the strongest RAE recipe, XRAE improves FDr$^{6}$ in both the guided and non-guided settings and reaches a near-state-of-the-art non-guided gFID, demonstrating that exploration helps even the strongest, most heavily-tuned recipes at scale. We rely primarily on FDr$^{6}$, which has started to become standard in this setting~\cite{yang2026representation, singh2026improved}, as gFID is highly saturated and often \textit{misrepresents} sample quality~\cite{yang2026representation}---so XRAE's slightly worse guided gFID likely reflects this saturation.
Table adapted from~\cite{zheng2025representation}; we report the RAE baseline performance under our setup.}
\label{tab:rae_comparison}
\vspace{-15pt}
\end{table}

\begin{takeaway}
\textbf{Takeaway:} Exploration improves even state-of-the-art recipes' sample efficiency by more than $6\times$, reaching a near-state-of-the-art unguided FID, and converging almost $300\times$ faster than the standard SiT recipe.
\end{takeaway}

\paragraph{Do Improvements From Exploration Vary With Scale?}

Throughout our experiments, exploration has helped more at larger scale---most notably, its efficiency gains more than doubled when moving from the SiT setting to the RAE setting, which used $3\times$ the compute.
This follows from how a generative model's performance is limited by three capacities: parameters restrict what it can represent, data restricts what it can learn, and generative expressivity restricts what it can generate. Conventional scaling of parameters and data relieves those constraints, but generative expressivity is set by the training objective itself (Section~\ref{sec:mode_forcing}; see Section~\ref{sec:generative_expressivity_details} for the formal scope), so it stays fixed regardless of how large models or datasets grow.
At small scale, this fixed generative expressivity is generally not an issue, as models are primarily held back by limited parameters and data. However, as parameter and data scale increase, generative expressivity increasingly becomes the bottleneck. Therefore, since exploration raises generative expressivity directly, we hypothesize its benefits should grow as models and data scale.

\begin{figure}[t]
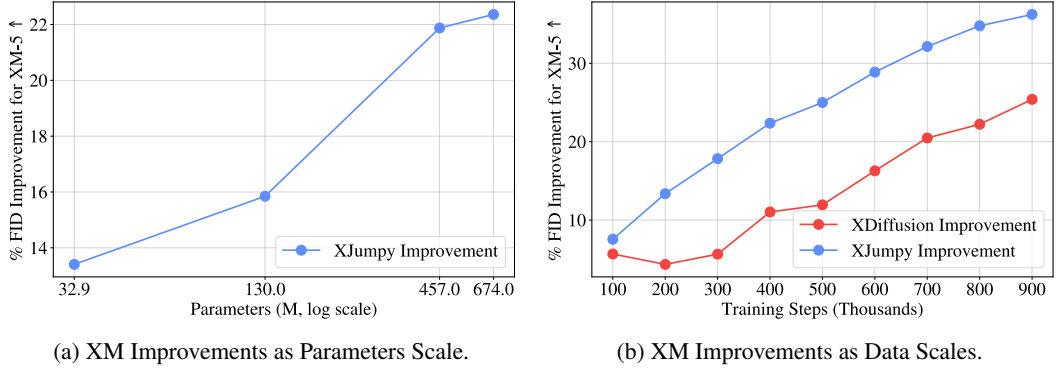

  \centering
  \begin{subfigure}[b]{0.49\columnwidth}
    \includesvg[width=\linewidth]{fig/xm_param_pct_improvement_jumpy.svg}
    \caption{XM Improvements as Parameters Scale.}
    \label{subfig:xm_param_gains_scale}
  \end{subfigure}\hfill
  \begin{subfigure}[b]{0.49\columnwidth}
    \includesvg[width=\linewidth]{fig/xm_steps_pct_improvement_overlap.svg}
    \caption{XM Improvements as Data Scales.}
    \label{subfig:xm_data_gains_scale}
  \end{subfigure}
  \caption{\textbf{Performance Gains from Exploration Increase as Scale Increases.} We measure the performance gain from exploring 5 modes (XM-5) over no exploration as we scale model size (left) and training data (right). In both cases the gains from doing exploration grow with scale, rising from $13\%$ to $23\%$ with model size and from $7\%$ to $36\%$ with data.}
  \label{fig:xm_data_param_gains_scale}
  \vspace{-15pt}
\end{figure}

To test this, we measure the gains from exploration while varying model size and data scale, and find that gains rise from $7\%$ to $36\%$ as data scales, and from $13\%$ to $23\%$ as model size scales (Figure~\ref{fig:xm_data_param_gains_scale}).
The FLOP and sample efficiency experiments reinforce this---exploration helped more the longer a model was trained, with the FLOP-optimal amount of exploration growing over the course of training (Figures~\ref{subfig:xrae_flop_eff} and~\ref{subfig:fid_xm_flops_scaling}).
Together, these results point to exploration as a \textit{missing scaling axis in existing generative models}, where the compute-optimal amount of exploration grows with scale just like parameters and data. This means today's generative models, trained without exploration, \textit{increasingly fall short} of what \textit{compute-optimal exploration} would achieve.
Foundation-model training runs use roughly four orders of magnitude more compute than our largest experiments, so if this trend continues, the improvements reported here likely underestimate the gains at that scale.

\begin{takeaway}
\textbf{Takeaway:} Gains from exploration grow with scale---as models and data scale, the bottleneck increasingly becomes generative expressivity, which exploration raises directly.
\end{takeaway}

\paragraph{Which Models Benefit Most From Exploration?} So far, exploration has helped every model family we tested, raising a natural question---are some generative models better suited to exploration than others?

To investigate this, we compare Diffusion/Flow with Jumpy generative models. Jumpy models work by interpolating between direct end-to-end regression ($1$ jump) and continuous-time Flow ($\infty$ jumps), so using a finite number of jumps is more end-to-end than Flow and more \textit{generatively expressive} than a single-step regressor (More details in Section~\ref{sec:existing_gen_models}).

We compare XDiffusion and XJumpy generative models for FID and FVD scaling as exploration increases in Figure~\ref{fig:xm_explore_fid_fvd_scaling}, where the rate of improvement for XJumpy models as exploration increases is much higher than the rate for XDiffusion. For example, in Figure~\ref{subfig:fid_xm_explore_scaling}, XJumpy generative models start out as performing worse than XDiffusion models with no exploration, but as exploration increases XJumpy models become more performant than XDiffusion models. 
This is further reinforced by Figure~\ref{subfig:xm_data_gains_scale}, where XJumpy models see larger gains from exploration than XDiffusion as data scales. 
Together, these results suggest that generative models that are more end-to-end scale better with increased exploration.

\begin{wrapfigure}{r}{0.55\textwidth}
    \centering
    \includesvg[width=\linewidth]{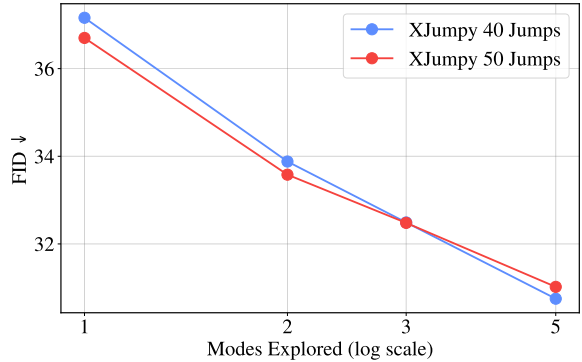}
    \caption{\textbf{More End-to-End Models Scale Better with Exploration.} We compare Explorative Jumpy (XJumpy) models with a different number of jumps across an increasing number of modes explored. If the amount of exploration is low, XJumpy models with more jumps perform best. However, as exploration increases, the optimal number of jumps decreases, demonstrating how models that are more end-to-end (fewer jumps) scale better with increased exploration.}
    \label{fig:xm_end_to_end_scaling}
    \vspace{-15pt}
\end{wrapfigure}

We can test this hypothesis further by varying the number of jumps within an XJumpy model. 
If factoring training can substitute for factoring generation, then more exploration should decrease the optimal number of jumps, since exploration supplies the generative expressivity those extra jumps would otherwise provide.
Figure~\ref{fig:xm_end_to_end_scaling} shows this, where an XJumpy model with fewer jumps scales better with exploration than an XJumpy model with more jumps. These results directly demonstrate that as exploration increases, more end-to-end models perform better---in effect, \textit{exploration scales how end-to-end existing generative models can be}. This makes Jumpy models, which can be more end-to-end than Diffusion/Flow, a promising approach to pair with exploration, and offers an indirect route to better generalization, as more end-to-end models reduce exposure bias (Section~\ref{sec:e2e_gen_modeling}).

\begin{takeaway}
\textbf{Takeaway:} Models that are \textit{more end-to-end} benefit most from exploration, so as increased exploration supplies needed generative expressivity, the best-performing models become increasingly end-to-end---turning how end-to-end models are from a fixed design choice into a scalable one.
\end{takeaway}

\subsection{End-to-End Explorative Models}
\label{sec:xm_e2e}

\paragraph{Can Explorative Modeling be Used for End-to-End Generation?} So far, exploration has been combined with existing generative models, where we found it enables them to become more end-to-end (Figure~\ref{fig:xm_end_to_end_scaling}); we now take this trend to its limit, using Explorative Models as \textit{standalone} end-to-end generative models, where sampling is identical at training and inference (Section~\ref{sec:e2e_gen_modeling}). We evaluate end-to-end XMs on robotics control tasks, including Behavior Cloning, comparing our Explorative Policy to Diffusion Policy~\cite{chi2023diffusion} (Table~\ref{tab:bc_results}), and Goal-Conditioned World Modeling, comparing our Explorative World Model to Diffuser~\cite{janner2022planning} (Table~\ref{tab:gcwm_results}). 

In both settings, Explorative Models match diffusion baselines at a fraction of the inference compute---Explorative Policy rivals Diffusion Policy with a single network forward pass instead of 100, and the Explorative World Model matches Diffuser using $16$-$256\times$ fewer function evaluations. This gap comes directly from what each approach factors: diffusion pays for its generative expressivity with generation steps at inference, while end-to-end XMs pay for it with exploration during training, keeping inference at a single forward pass. Notably, we obtain these results with barely any tuning of hyperparameters for XMs---we keep each baseline's architecture and occasionally add a recurrent block for an inductive bias toward recurrence---so we believe these results underrepresent how well-tuned XMs can perform with additional tricks. The main limitation of these experiments is in handling highly multimodal distributions. Because we use Forward XM here, which has a cost that grows with the number of modes explored, it cannot cheaply cover extremely multimodal distributions. Therefore, Reverse XM is likely better suited for end-to-end generation tasks, which we largely leave for future work (we discuss successful Reverse XM training in Section~\ref{sec:additional_experiments}, and future directions in Section~\ref{sec:future_work}).

\begin{table}[t]
\centering
\small
\renewcommand{\arraystretch}{1.2}
\setlength{\tabcolsep}{8pt}
\begin{tabular}{l c ccccc}
\toprule
\textbf{Method} & \textbf{NFE}$\downarrow$ & \textbf{Lift}$\uparrow$ & \textbf{Can}$\uparrow$ & \textbf{Square}$\uparrow$ & \textbf{Transport}$\uparrow$ & \textbf{Tool Hang}$\uparrow$ \\
\midrule
\multicolumn{7}{l}{{\textit{Proficient Human, State Observations}}} \\
\arrayrulecolor{black!30}\midrule\arrayrulecolor{black}
Diffusion Policy~\cite{chi2023diffusion} & 100 & $\mathbf{100\%}$ & $\mathbf{100\%}$ & $94\%$ & $72\%$ & $\mathbf{86\%}$ \\
\textbf{Explorative Policy} & \textbf{1} & $\mathbf{100\%}$ & $\mathbf{100\%}$ & $\mathbf{96\%}$ & $\mathbf{74\%}$ & $\mathbf{86\%}$ \\
\bottomrule
\end{tabular}
\caption{\textbf{Explorative Policy Rivals Diffusion Policy at $\mathbf{100\times}$ Less Inference Compute.} Following the setup of Diffusion Policy~\cite{chi2023diffusion}, we report Behavior Cloning success rates on Robomimic tasks under the proficient-human, state-observation setting. Explorative Policy, our end-to-end Explorative Modeling-based policy, takes a single network forward pass (NFE: 1) at inference, while Diffusion Policy requires 100. Despite using significantly less inference compute, Explorative Policy matches or surpasses Diffusion Policy on all benchmarks.}
\label{tab:bc_results}
\vspace{-5pt}
\end{table}

\begin{table}[t]
\centering
\small
\renewcommand{\arraystretch}{1.2}
\setlength{\tabcolsep}{6pt}
\resizebox{\columnwidth}{!}{
\begin{tabular}{l cc cc cc cc}
\toprule
\multirow{2}{*}{\textbf{Method}} & \multicolumn{2}{c}{\textbf{U-Maze}} & \multicolumn{2}{c}{\textbf{Medium}} & \multicolumn{2}{c}{\textbf{Large}} & \multicolumn{2}{c}{\textbf{Average}} \\
\cmidrule(lr){2-3} \cmidrule(lr){4-5} \cmidrule(lr){6-7} \cmidrule(lr){8-9}
 & Score$\uparrow$ & NFE$\downarrow$ & Score$\uparrow$ & NFE$\downarrow$ & Score$\uparrow$ & NFE$\downarrow$ & Score$\uparrow$ & NFE$\downarrow$ \\
\midrule
\multicolumn{9}{l}{{\textit{Maze2D, Goal-Conditioned}}} \\
\arrayrulecolor{black!30}\midrule\arrayrulecolor{black}
Diffuser~\cite{janner2022planning} & $118.7$ & $64$ & $\mathbf{128.5}$ & $256$ & $134.4$ & $256$ & $127.2$ & $192$ \\
\textbf{Explorative World Model} & $\mathbf{121.4}$ & \textbf{4} & $122.9$ & \textbf{1} & $\mathbf{145.8}$ & \textbf{1.9} & $\mathbf{130.0}$ & \textbf{2.3} \\
\bottomrule
\end{tabular}
}
\caption{\textbf{Explorative World Model Matches Diffuser While Using $\mathbf{16-256\times}$ Less Inference Compute.} Goal-conditioned world modeling performance on the Maze2D tasks~\cite{fu2020d4rl, janner2022planning}. Our Explorative World Model is compared against Diffuser~\cite{janner2022planning}, and achieves better average performance while using $80\times$ less inference compute on average. Some Explorative World Models take more than one NFE due to recurrent blocks~\cite{geiping2026scaling}.}
\label{tab:gcwm_results}
\vspace{-15pt}
\end{table}

\begin{takeaway}
\textbf{Takeaway:} Explorative Modeling enables scalable end-to-end reconstructive generative models, matching strong diffusion baselines on robotics and world modeling tasks while using up to $256\times$ less inference compute.
\end{takeaway}

%% file: main_sec/6_discussion.tex
\section{Discussion}
\label{sec:discussion}

\paragraph{Mode Forcing as a Predictive Theory.}
Much of deep learning progresses by running experiments first and explaining them afterward.
Because this work builds on Mode Forcing~\cite{gladstone2026mode}, most of its results came about in the opposite order, where the theory predicted them before the experiments were run. Here we outline Mode Forcing's predictions and their confirmations:
\begin{itemize}
    \item \textbf{Even the strongest generative models are short on generative expressivity.} Mode Forcing argues that factoring generation often still leaves modes uncaptured, with heavy reliance on guidance as evidence for this (Section~\ref{sec:mode_forcing}). Added generative expressivity should therefore improve even the most heavily tuned recipes, and it does: exploration lifts image, video, and language generation performance for all recipes tested.
    
    \item \textbf{Generative expressivity increasingly becomes the bottleneck at scale.} As parameters and data stop limiting what models can represent and learn, Mode Forcing predicts that generative expressivity set by the training objective should increasingly become the bottleneck. We find exactly this, where gains from exploration climb from $7\%$ to $36\%$ as data scales and $13\%$ to $23\%$ as models grow (Figure~\ref{fig:xm_data_param_gains_scale}), and the FLOP-optimal amount of exploration rises over the course of training (Figures~\ref{subfig:xrae_flop_eff} and~\ref{subfig:fid_xm_flops_scaling}).

    \item \textbf{Exploration can substitute for generation factorization.} As discussed in Section~\ref{sec:mode_forcing}, factoring generation exists to supply generative expressivity, so supplying it through exploration instead should reduce how much generation factorization a model needs. We confirm this, where as exploration grows, the optimal amount of generation factorization decreases, and more end-to-end models perform better (Figure~\ref{fig:xm_end_to_end_scaling}). 
    
    \item \textbf{XMs enable end-to-end generation.} Mode Forcing argues that multimodal distributions are the core reason generation has to be factored, so if exploration handles multimodal distributions during training, end-to-end reconstruction should work. Our Explorative Policy and World Model confirm this, matching diffusion baselines with as little as a single forward pass instead of hundreds (Tables~\ref{tab:bc_results} and~\ref{tab:gcwm_results}).
\end{itemize}

\paragraph{The Benefits of Surplus Generative Expressivity.}
Interestingly, exploration helps even when generative expressivity is not a large bottleneck. In our video generation experiments, XMs improve performance significantly even though the modeled distribution is not very multimodal---Jumpy models need only 10 steps in this setting (Figure~\ref{fig:xm_explore_fid_fvd_scaling}), far less generation factorization than modern Diffusion models. Why would exploration help when there are few modes to capture? Our hypothesis is that even a weakly multimodal target still pulls each prediction toward multiple competing values over the course of training, and the same conflicting pulls that blur modes also make optimization harder. Exploration relieves this pressure, letting each prediction train toward its nearest match, so targets compromise less and optimization becomes easier. This mirrors overparametrization, where models with far more parameters than needed to fit their data consistently perform and generalize better~\cite{kim2025pretraining, slowrun_2026}---commonly attributed to smoother loss landscapes and a bias toward simpler solutions~\cite{li2018visualizing, wilson2025deep}. In both cases, surplus capacity makes good solutions easier to find, suggesting that exploration, like parameters, is worth scaling past the point where it seems strictly necessary.

%% file: main_sec/7_broader_impact_future_work.tex
\section{Future Works and Broader Impact}
\label{sec:future_work}

XMs open several research directions, below we highlight some of these directions.

\vspace{-5pt}
\paragraph{Exploration as a Scaling Axis for More Generative Models.} We demonstrated exploration acts as a scaling axis for Diffusion/Flow, Jumpy, and masked diffusion language models; we believe other generative models likely benefit from exploration in the same manner.
Autoregressive LLMs have proven the hardest case, for the reasons discussed in our limitations (Section~\ref{sec:conclusion}).
Evaluation is also part of the challenge, as language modeling lacks robust distributional metrics like FID and FVD that would reveal mode coverage. We see two promising paths to build on these early gains. Multi-token prediction~\cite{gloeckle2024better} targets are more multimodal, so multi-token prediction suffers more from limited generative expressivity and gives exploration more to offer. The Free Transformer~\cite{fleuret2025freetransformer} conditions a decoder on a latent variable inferred by a VAE, which is exactly the kind of latent exploration searches over, and training it with Explorative Modeling instead would remove the VAE entirely, along with the exposure bias of training on inferred latents (Section~\ref{sec:intuition}). Beyond language models, few-step models such as MeanFlow~\cite{geng2026mean} are a natural fit for exploration as well, since exploration can supply the generative expressivity their shortened trajectories amortize. We also believe in combining XMs with Energy-Based Transformers (EBTs)~\cite{gladstone2025energy}, where the biggest documented challenge with EBTs has been end-to-end generation and handling highly multimodal distributions, which is exactly what XMs enable. Paired together, XMs and EBTs could enable more dynamic reasoning, search, and generalization over entire sequences.

\vspace{-5pt}
\paragraph{End-to-end XM Applications.}
End-to-end XMs are especially well suited to new applications such as inpainting and super-resolution due to low amounts of multimodality in generated distributions.
They could also pair with feature-based world models like JEPA~\cite{assran2023selfsupervised} to build end-to-end world models. In the short term this pairing is especially practical for Forward XM, as feature spaces often contain far fewer modes than raw observations~\cite{lecun2022path}, so a small $K$ suffices (in the long run, we believe XMs can scale to arbitrarily multimodal settings via Reverse XM, Section~\ref{sec:additional_experiments}). Exploration would also resolve a core JEPA challenge: next-state prediction and trajectory-level planning are multimodal, especially in non-deterministic environments, which feature regression blurs but exploration captures.
Another appealing direction is combining exploration with moment matching~\cite{li2015generative}, which would let models reconstruct features at the right granularity. 
Finally, end-to-end XMs further enable setting the number of modes a model captures, which existing generative models struggle with, and because Forward XM favors recall while Reverse XM favors precision, choosing between or combining them gives direct control over generation diversity.

\vspace{-5pt}
\paragraph{Improving and Understanding XMs.} There is plenty of room to improve the core mechanism of exploration itself; in this paper we primarily used the simplest approach of sampling many different random noise candidates for Diffusion/Flow/Jumpy models. In principle, architectures could condition on discrete latent embeddings for each explorative factor, which could give better controllability, enable more uniform sampling of modes, and improve mode coverage (we did this for MDLMs, but no other models). Additionally, there are likely better exploration approaches that exist. In this work, exploration was done by drawing $K$ independent candidates. In principle, however, searching for the optimal latent could be done in better ways, such as by treating the reconstruction loss as an energy, and finding the best latent by gradient descent~\cite{du2022learning}. The risk in using this approach is that it could cause a mismatch reminiscent of VAE prior holes~\cite{rosca2018distribution}, where the latents found by search differ from those sampled at inference, though applying such search only late in training or using other tricks could avoid this.
Forward XM's cost could also be cut with a cheaper scorer, such as a smaller proxy that ranks candidates so only the winner is generated in full. Training on the soft min rather than the hard min is another variant, letting every candidate contribute gradients and carrying the cleaner maximum likelihood interpretation (Section~\ref{sec:xm_kl_theory}).
Another interesting idea would be to unify exploration with an end-to-end learned encoder, so search happens over learned latents (this could be combined with recent work on learning generative models and encoders jointly, such as Unified Latents~\cite{heek2026unified}).
Finally, exploration deserves the same scaling-law treatment as parameters and data: the compute-optimal amount of exploration already grows with scale (Figures~\ref{subfig:xrae_flop_eff} and~\ref{subfig:fid_xm_flops_scaling}), so understanding how to optimally allocate compute between exploration, parameters, and data, similar to Chinchilla~\cite{hoffmann2022training}, would be insightful.

\vspace{-5pt}
\paragraph{Scaling Reverse XMs.}
Despite Reverse XMs' potential to collapse (Section~\ref{sec:forward_reverse_xm}), we see them as more promising in the long run over Forward XMs, as they add almost no extra FLOPs and scale more gracefully with the number of modes. With discrete conditioning, Reverse XMs come essentially for free, since loading a larger batch with $K$ data points per condition lets each generation pick its best match (we could have done this for our image generation experiments, but chose not to in order to keep implementations simple and modality/domain agnostic). 
Doing Reverse XM with continuous conditioning is harder, as data points rarely share the exact same condition, so each generation has no ready-made set of valid targets to search, and the central design question becomes how data is loaded. More ambitiously, a vector database over the whole dataset would let each generation search all training data in logarithmic time, so the number of modes explored can in principle reach the dataset size, directly matching the generated distribution to the training distribution. We have already made Reverse XMs work this way on language modeling tasks (more on this in Section~\ref{sec:additional_experiments}). One remaining challenge is that a generation's nearest datapoint can flip-flop across training steps, blurring the effective target; \textit{sticky} couplings that persist matches across steps could prevent this.

\paragraph{Exploration beyond Pretraining.}
The mode collapse XMs address during pretraining also often shows up in post-training, where RL fine-tuning is known to sharpen models onto a narrow set of behaviors~\cite{anthony2025kl}.
Recent fixes such as pass@$k$ rewards~\cite{chen2025passk} and best-of-$N$-aware fine-tuning~\cite{chow2024inference} can be seen through our lens as Forward XM, with a verifier standing in for ground truth data. These fixes act only during post-training, though; pretraining with exploration may yield base models that capture more modes in the first place, leaving RL more to select among.

%% file: main_sec/8_conclusion_limitations.tex
\section{Limitations and Conclusion}
\label{sec:conclusion}

In this work, we introduced Explorative Modeling, a new paradigm for handling multimodal distributions that factors the training loop instead of the generation procedure. Exploration increases generative expressivity, adding a new pretraining axis for existing generative models, and enabling end-to-end generative modeling.

\vspace{-5pt}
\paragraph{Limitations.}
As a scaling axis for existing models, exploration is easier to integrate into some model families than others. This is because exploration requires a latent variable to search over when selecting the best of $K$ candidates. Continuous generative models benefitted most easily, as they already condition on a noise $z$; MDLMs benefited once given a learned latent variable embedding for exploration. 
We found autoregressive language models harder to improve with exploration, likely because injecting a latent into them is less natural, and because they are less bottlenecked by generative expressivity than many other models. Despite this, we have achieved initial modest results showing improved data efficiency, suggesting exploration can benefit autoregressive LLMs further with more effort.

Exploration also changes the training objective, so losses are no longer directly comparable across exploration levels. This makes distributional metrics such as FID and FVD, as well as downstream metrics such as accuracy, more important for evaluation.
Similarly, existing guidance techniques were designed without exploration in mind, and some transferred to XMs better than others: autoguidance worked decently, while classifier-free guidance helped less than it does for base models, despite XMs' stronger unguided FID indicating they capture the underlying density better. Guidance is known to not transfer uniformly across models---for example, vanilla CFG also fails to improve models trained on representation autoencoder latents by default~\cite{zheng2025representation}---suggesting even our autoguidance results likely undershoot what XMs could achieve with guidance designed for exploration. Exploration also supplies a signal base models lack, namely $K$ candidates and a notion of which was best, so we believe using it to design guidance tailored specifically to XMs is one of the most important open problems.

Fully end-to-end XMs face a couple of challenges. The most significant challenge with end-to-end XMs is in handling highly multimodal data: fully end-to-end Forward XMs need $K$ to grow with the number of modes, which is currently too expensive for distributions with very many modes (e.g., image generation). Therefore, while the world remains somewhat compute constrained, Reverse XMs are a natural solution for handling high distribution multimodality, where a single model generation can search arbitrarily many training data points.
Reverse XMs do bring their own considerations: being mode-seeking, they require an entropy term or coverage constraint to avoid collapse, and searching data efficiently requires good representations along with a dataloader or vector database supporting the search.
Another challenge is that end-to-end XMs give up the implicit regularization of factored generation: when each step trains on corrupted or partial inputs, memorization is harder, so fully end-to-end models are more prone to memorizing when data is scarce. This concern fades with scale, however, as having abundant data itself acts as regularization~\cite{somepalli2022diffusion, kadkhodaie2024generalization, gu2023memorization}. Taken to the limit, with enough data and compute, it's plausible that generative modeling simply becomes exploration for good latents using a very high $K$.

\vspace{-5pt}
\paragraph{Conclusion.} Across both continuous and discrete domains, we found exploration acts as a third pretraining axis alongside parameters and data, with gains that \textit{grow with scale} rather than saturate---rising from $7\%$ to $36\%$ as data scales, $13\%$ to $23\%$ as models grow, and with efficiency gains more than doubling at $3\times$ the compute. Because gains from exploration keep climbing with scale, the numbers we report are likely a floor for benefits at increased model scale. Concretely, exploration improves FLOP efficiency by $4.1\times$, sample efficiency by $6.2\times$, and parameter efficiency by $47\%$, while lifting the strongest of image-generation recipes to a near-state-of-the-art $1.43$ FID on ImageNet without guidance. Beyond efficiency, exploration enables \textit{scaling generalization}: spending more training compute on exploration improves generalization directly, and improves it indirectly by enabling existing models to become more end-to-end. Taken to its limit, exploration enables fully end-to-end reconstructive generation, matching diffusion on control tasks with as little as a single forward pass in place of hundreds.
For over a decade \textit{we have scaled how large generative models are and how much data they train on}; XMs let us scale \textit{what models can generate}.

%% file: main_sec/9_acknowledgements.tex
\section*{Author Contributions}

\vspace{-5pt}
\paragraph{Alexi Gladstone} led the project from ideation to execution, conceiving Explorative Modeling, developing the theory and method, designing and running all experiments, and writing the paper.

\vspace{-5pt}
\paragraph{Heng Ji and Yilun Du} advised the project throughout, providing invaluable mentorship, feedback on the ideas and writing, and compute support. Yilun had crucial initial ideas on XMs for training EBMs.

\section*{Acknowledgement}
Huge thanks to Flapping Airplanes for supporting Alexi as a fellow while completing this work. Massive thanks to Laude Institute for supporting this work, with a special shoutout to Braden Hancock and K. Tighe. Thanks to Soran Ghaderi for productive early discussions related to XMs. Thanks to Omead Pooladzandi and Samip Dahal for great feedback on XMs.
This material is based upon work supported by the U.S. National Science Foundation Graduate Research Fellowship Program under Grant No. DGE 21-46756, U.S. DARPA ECOLE Program No. \#HR00112390060, DARPA ITM Program No. FA8650-23-C-7316, NSF Molecule Maker Lab Institute, an AI Institute for Molecular Discovery, Synthesis Strategy, and Manufacturing funded by the U.S. National Science Foundation under Awards No. 2019897 and 2505932, the AI Research Institutes program by National Science Foundation and the Institute of Education Sciences, U.S. Department of Education through Award No. 2229873 - AI Institute for Transforming Education for Children with Speech and Language Processing Challenges, and NSF NAIRR award. 
Any opinions, findings, and conclusions or recommendations expressed in this material are those of the author(s) and do not necessarily reflect the views of the National Science Foundation, the Defense Advanced Research Projects Agency (DARPA), the Institute of Education Sciences, or the U.S. Department of Education.
This research used the Delta and DeltaAI advanced computing and data resources, which are supported by the National Science Foundation (award OAC 2320345 and award OAC 2005572) and the State of Illinois. Delta and DeltaAI are joint efforts of the University of Illinois Urbana-Champaign and its National Center for Supercomputing Applications.
Some of the computations in this paper were run on the FASRC cluster supported by the FAS Division of Science Research Computing Group at Harvard University.

%% file: supp_sec/A_additional_experiments.tex
\section{Additional Experimentation}
\label{sec:additional_experiments}

\begin{figure}[t]
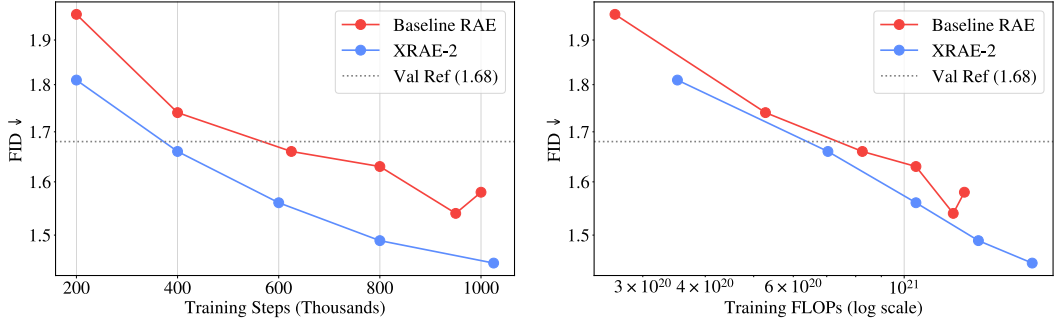

  \centering
  \begin{subfigure}[b]{0.49\columnwidth}
    \includesvg[width=\linewidth]{fig/xm_rae_fid_scaling_ema.svg}
  \end{subfigure}\hfill
  \begin{subfigure}[b]{0.49\columnwidth}
    \includesvg[width=\linewidth]{fig/xm_rae_fid_scaling_ema_flops.svg}
  \end{subfigure}
  \caption{\textbf{RAE FID Convergence.} The same comparison as Figure~\ref{fig:xm_rae_flop_step_eff} measured with FID rather than $\mathrm{FDr}^{6}$: XRAE converges much faster in terms of FLOPs and training steps than the baseline RAE. The improvement gap is higher for $\mathrm{FDr}^{6}$, as FID at this level of performance is saturated and no longer tracks true sample quality~\cite{yang2026representation, singh2026improved}.}
  \label{fig:xm_rae_fid_scaling}
\end{figure}

Figure~\ref{fig:xm_rae_fid_scaling} reports the FID versions of the $\mathrm{FDr}^{6}$ convergence plots in Figure~\ref{fig:xm_rae_flop_step_eff}, where XRAE similarly converges much faster than the baseline.

Though we do not report them in detail, we have also observed these benefits in the discrete domain, where adding exploration to MDLMs improves both sample efficiency and generalization, mirroring the trends for image and video generation (Figures~\ref{fig:xm_rae_flop_step_eff} and~\ref{fig:fvd_xm_overfitting}).
Similarly, we have observed early evidence that exploration improves the data efficiency of autoregressive language models, which we leave to future work to report in detail.

\subsection{Exact Results for Exploration Scaling}
\label{sec:explore_scaling_values}

For reproducibility, and to make future comparisons easier, Table~\ref{tab:xm_explore_scaling_values} reports the exact FID and FVD values behind Figure~\ref{fig:xm_explore_fid_fvd_scaling}.

\begin{table}[h]
\centering
\small
\renewcommand{\arraystretch}{1.1}
\setlength{\tabcolsep}{6pt}
\begin{tabular}{l ccccccc}
\toprule
 & \multicolumn{7}{c}{\textbf{Modes Explored (XM-$K$)}} \\
\cmidrule(lr){2-8}
\textbf{Model} & \textbf{1} & \textbf{2} & \textbf{3} & \textbf{5} & \textbf{8} & \textbf{12} & \textbf{25} \\
\midrule
\multicolumn{8}{l}{\textit{FID$\downarrow$, ImageNet 256$\times$256, Small models (Figure~\ref{subfig:fid_xm_explore_scaling})}} \\
\arrayrulecolor{black!30}\midrule\arrayrulecolor{black}
XDiffusion 50 Steps & 62.5 & 57.8 & 57.1 & 56.2 & -- & 55.0 & 54.6 \\
XJumpy 50 Steps & 63.3 & 59.1 & 57.5 & 55.6 & -- & 53.8 & \textbf{53.2} \\
\midrule
\multicolumn{8}{l}{\textit{FVD$\downarrow$, Something-Something V2, Base models (Figure~\ref{subfig:fvd_xm_explore_scaling})}} \\
\arrayrulecolor{black!30}\midrule\arrayrulecolor{black}
XDiffusion 50 Steps & 36.9 & 33.3 & 32.3 & 31.3 & 30.6 & 30.0 & -- \\
XJumpy 10 Steps & 26.9 & 24.0 & 23.2 & 21.6 & \textbf{21.2} & -- & -- \\
\bottomrule
\end{tabular}
\caption{\textbf{Exact FID and FVD Values Across Exploration Scaling.} The values underlying Figure~\ref{fig:xm_explore_fid_fvd_scaling}, reported to ease reproduction and comparison in future work. Dashes mark exploration levels not tested for that setting.}
\label{tab:xm_explore_scaling_values}
\end{table}

\subsection{Comparing Exploration to Minibatch Optimal Transport Couplings}
\label{sec:ot_couplings}

A natural alternative to exploration is to reduce mode blurring by computing a better coupling directly, most commonly with minibatch Optimal Transport (OT)~\cite{tong2023improving, pooladian2023multisample}. Comparing the two, we find minibatch OT couplings actually \textit{hurt} image generation performance, worsening FID from $46.3$ to $54.5$ for Small models and from $74.4$ to $82.7$ for Base models (Table~\ref{tab:ot_comparison}).

We attribute this to two problems that exploration avoids. First, minibatch OT is a biased approximation of the global coupling~\cite{fatras2020learning}, and this bias worsens as datasets grow, since each batch covers a vanishing fraction of the data. Second, the two select couplings on different grounds. Minibatch OT assigns pairs by geometry alone, computed within a batch regardless of what the model has learned; exploration selects by the model's own current loss, so its coupling co-adapts with the model throughout training. Empirically, the model-aligned choice helps while the geometric one hurts, and exploration's gains \textit{grow} with scale rather than degrade (Figure~\ref{fig:xm_data_param_gains_scale}).

\begin{table}[h]
\centering
\small
\renewcommand{\arraystretch}{1.1}
\begin{tabular}{l cc}
\toprule
\textbf{Method} & \textbf{Small FID}$\downarrow$ & \textbf{Base FID}$\downarrow$ \\
\midrule
Flow Matching & \textbf{46.3} & \textbf{74.4} \\
Flow Matching + Minibatch OT & 54.5 & 82.7 \\
\bottomrule
\end{tabular}
\caption{\textbf{Minibatch OT Couplings Hurt Performance.} FID after $200$k training steps for Flow Matching models trained with and without minibatch OT couplings. OT hurts performance at both model sizes, which we attribute to the bias of minibatch couplings and their model-agnostic assignment. The setup here is ImageNet-1k~\cite{russakovsky2015imagenet} class-conditional image generation.}
\label{tab:ot_comparison}
\end{table}

\subsection{Reverse XM Language Models}
\label{sec:reverse_xm_lms}

We have successfully trained Reverse XM language models, where each generation searches the training data through a vector database rather than comparing against $K$ samples in a batch. Getting this to work relied on two ingredients. First, once a data point is matched and trained on, it is removed from the search pool for that epoch, which we track as a \textit{train coverage percent}. This prevents nearest-neighbor search from collapsing onto the same few points, playing the role of the entropy term Reverse XM needs to avoid collapse (Section~\ref{sec:xm_kl_theory}), though as a coverage heuristic rather than a literal entropy bonus. Additionally, we found it important to set the \textit{train coverage percent} to be less than the entire train set (e.g., $50\%$), or else models have high loss near the end of an epoch as they try to cover the extremes of a dataset. Second, the search happens in a hybrid of representation space and cross-entropy space, so that retrieved neighbors reflect the actual training loss. We strongly believe this can be further improved upon, and we have not yet open sourced this code, but plan to.

%% file: supp_sec/B_Additional_Intuition.tex
\section{Additional Intuition}

\paragraph{Generative Modeling with a For Loop.}
It is worth appreciating how simple end-to-end Explorative Modeling can be. Unlike diffusion, flow, or autoregressive models, which rely on many-step sampling procedures at inference to avoid mode blurring and often complicated masking/noising schedules (Section~\ref{sec:mode_forcing}), end-to-end XMs generate with a single pass from the model and training XMs in the simplest case is just a short for loop with 3-5 lines of code (Algorithms~\ref{alg:min_samples} and~\ref{alg:min_data}).

\paragraph{Exploration as Building an Associative Memory.}
A generative model turns an input, usually noise drawn from a multivariate Gaussian, into a sample. Explorative Modeling searches over this noise to find which inputs the model should tie to which samples. Once trained, each region of noise acts as a key that retrieves one mode of the data as its value, so XMs behave much like associative memories. A larger $K$ splits the noise into finer regions, so each mode gets its own instead of blurring together (Figure~\ref{fig:exploration_improves_coverage}). This holds for every generative model, not just XMs. However, it \textit{does not mean} that generative models simply memorize the training data. Large datasets act as a form of regularization: with far more examples than the model can store individually, it is forced to reuse parameters across them and learn the structure they share instead of the datapoints themselves, which results in generalization.

%% file: supp_sec/C_Approach_Details.tex
\section{Approach Details}
\label{sec:approach_details}

\paragraph{Batching Forward XM.}
Algorithm~\ref{alg:min_samples} is written as a for loop for clarity, but in practice the $K$ explored generations can be folded into the batch dimension and computed as a single larger forward pass. Because accelerators process larger batches efficiently, this parallelizes exploration and speeds up training considerably compared to looping over candidates one at a time. We provide batched code for Forward XM in the released code.

\paragraph{Other Ways to Explore.}
Drawing $K$ fresh candidates per step is not the only way to do Forward XM. For example, generations could be cached by their class or conditioning and reused as candidates for later samples with the same condition---on ImageNet, caching recently generated samples for each class would provide candidates essentially for free---though this works worse for continuous conditioning, where datapoints rarely share the exact same condition. We use fresh draws throughout, as this is the simplest approach to implement, is fairest when comparing across domains, and is less prone to collapse.

\paragraph{Gradients and Memory.}
Only the best candidate receives gradients (Equation~\ref{eq:forward}), which can be implemented in two ways. In the \textit{memory-saving} mode, all $K$ candidates are forwarded without gradients, and only the best is re-forwarded with gradients to train on, keeping activation memory the same as standard training at the cost of one extra forward pass. In the \textit{FLOP-efficient} mode, all $K$ candidates are forwarded with gradients and only the lowest loss is backpropagated, avoiding the extra forward pass at the cost of storing activations for all $K$ candidates. Switching between the two lets exploration adapt to whichever of FLOPs or memory is the bottleneck.

\paragraph{Exploration FLOP Cost.}
For transformers, a training step costs roughly $6ND$ FLOPs for $N$ parameters and $D$ tokens, where the forward pass costs $2ND$ and the backward pass $4ND$~\cite{casson2023transformerflops}. Since only the best candidate is trained on, each additional explored candidate in Forward XM adds only a forward pass, so each additional mode explored costs roughly $\tfrac{1}{3}$ of a standard training step (XM-$K$ costs $\tfrac{K+2}{3}$ standard steps in the FLOP-efficient mode, plus one more forward pass in the memory-saving mode). Exploring more modes in Reverse XM is far cheaper, as additional data targets add no forward passes, only extra loss computations---essentially one matrix multiplication scaling with the data dimension---which is negligible compared to the network's FLOPs.

\paragraph{Generating the $K$ Candidates.}
For all continuous models in this work, both end-to-end XMs and the hybrid XMs built on Diffusion/Flow and Jumpy models (Section~\ref{sec:xm_scaling_axis}), each explored candidate uses a different input noise draw; for the hybrids, this means the same data sample, timestep, and condition (including condition dropping for guidance when the underlying recipe uses it), with only the noise varying. For XMDLMs, we instead learn $K$ discrete latent embeddings: each candidate samples one of these embeddings at random for every masked position, and the best-of-$K$ selection is over the resulting latent-conditioned predictions.

\paragraph{Implementing Reverse XM.}
Reverse XM requires each generation to have a set of valid data targets to search over. With discrete conditioning, this comes almost for free through the dataloader: loading $K$ datapoints per condition lets each generation pick its best match within the batch, at no extra generation cost. The same can be done with continuous conditioning, just involving additional tricks to load similar latents together. At larger scale, the whole dataset can instead be indexed in a vector database, letting each generation search all training data in roughly logarithmic time, so the number of modes explored can in principle reach the dataset size (we take this approach for the Reverse XM language modeling experiments described in Section~\ref{sec:additional_experiments}; see also Section~\ref{sec:future_work}).

%% file: supp_sec/D_Experimental_Details.tex
\section{Experimental Details}
\label{sec:experimental_details}

A lot of the experimentation paragraph and takeaway style is inspired by~\cite{sobal2026learning}. Some of the figure designs in this paper follow those of Mode Forcing~\cite{gladstone2026mode}. 

\paragraph{Model Sizes.}
All image and video generation models are transformers following the standard DiT size conventions~\cite{peebles2023scalable}, summarized in Table~\ref{tab:model_sizes}.

\begin{table}[h]
\centering
\small
\renewcommand{\arraystretch}{1.1}
\begin{tabular}{l ccc}
\toprule
\textbf{Size} & \textbf{Layers} & \textbf{Hidden Dim} & \textbf{Heads} \\
\midrule
Small  & 12 & 384  & 6  \\
Base  & 12 & 768  & 12 \\
Large  & 24 & 1024 & 16 \\
XLarge & 28 & 1152 & 16 \\
\bottomrule
\end{tabular}
\caption{\textbf{Model Sizes for Image and Video Generation.} Sizes follow the DiT conventions~\cite{peebles2023scalable}.}
\label{tab:model_sizes}
\end{table}

\paragraph{Image Generation.}
All image generation experiments train class-conditional models on ImageNet 256$\times$256. Aside from the RAE experiments, we follow the SiT setup~\cite{ma2024sit}, with the exception of not using a horizontal flip augmentation. Images are encoded by the VAE~\cite{stabilityai_sd_vae} into $4\times32\times32$ latents, which with a patch size of 2 gives a $16\times16$ grid of patches, or 256 tokens per image (this is the token count behind the data scaling in Figure~\ref{subfig:fid_xm_data_scaling}). All models train with a batch size of 256 and a learning rate of $1\mathrm{e}{-4}$, using 10k steps of linear warmup followed by cosine decay to 1M steps (or to 3M steps for the longer runs in Figure~\ref{subfig:fid_xm_flops_scaling}), along with gradient clipping of 1.0 and weight decay of 0.01. For Diffusion models we use the Flow Matching formulation, and following the RAE and JiT papers~\cite{zheng2025representation, li2026back} we sample with 50 Heun steps~\cite{karras2022elucidating} by default, which we found was enough to converge. For the FLOP scaling curves in Figure~\ref{subfig:fid_xm_flops_scaling} we use Jumpy models~\cite{gladstone2026jumpy}, as they were more stable for longer runs and more performant. These experiments all report validation-set generative metrics, as they better measure overfitting and generalization.

\paragraph{RAE Image Generation.}
For the RAE experiments (Table~\ref{tab:rae_comparison} and Figure~\ref{fig:xm_rae_flop_step_eff}) we follow the RAE setup~\cite{zheng2025representation}, building directly on their codebase and only adding exploration, including their batch size of 1024. We rely primarily on $\mathrm{FDr}^{6}$~\cite{yang2026representation} for these experiments, as FID at this level of performance is overfit and no longer tracks true sample quality~\cite{yang2026representation, singh2026improved}. For guided results, both the RAE baseline and XRAE use AutoGuidance~\cite{karras2024guiding}: the RAE baseline uses a guidance scale of 1.42 with the released S model at epoch 14 as the guiding model; for XRAE-2, the best $\mathrm{FDr}^{6}$ uses a scale of 1.5 with the same released S model at epoch 14, while the best gFID uses a scale of 1.35 with an XM-2 L model at epoch 9.

\paragraph{Video Generation.}
Video generation experiments use the Something-Something V2 dataset~\cite{goyal2017something} at $128\times128$ resolution, as higher resolutions required too much compute. We always model 10 frames, passing in frames 0, 1, and 9 as conditioning (simulating goal-conditioned world modeling~\cite{janner2022planning}), and use a 3D video transformer~\cite{assran2025vjepa} with Base model size and a patch size of 4, also chosen due to limited compute, as these experiments aim for fair comparisons rather than state-of-the-art results. Models train with a batch size of 256, a learning rate of $1\mathrm{e}{-4}$, and weight decay of $1\mathrm{e}{-2}$.

\paragraph{Individual Results in More Detail.}
Figure~\ref{subfig:fid_xm_explore_scaling} uses Small models, while Figure~\ref{subfig:fvd_xm_explore_scaling} uses Base models. Figure~\ref{fig:xm_flop_data_scaling} uses Base models for class-conditional image generation, where the baseline is the optimally tuned SiT recipe~\cite{ma2024sit}. Figure~\ref{fig:fvd_xm_overfitting} uses 4-step Jumpy models, plotting FVD over the course of training rather than the best performance reached. Figure~\ref{subfig:xm_param_gains_scale} measures image generation improvements across Jumpy model sizes from Small through XLarge, and Figure~\ref{subfig:xm_data_gains_scale} does the same for XLarge models across training steps. Figure~\ref{fig:xm_end_to_end_scaling} uses Base-sized image generation models.

\paragraph{Behavior Cloning.}
All Explorative Policy runs use XM-10. All Behavior Cloning models are CNNs rather than transformers, and we follow most of the Diffusion Policy setup~\cite{chi2023diffusion}. However, we use a newer version of robomimic~\cite{mandlekar2021matters}, so results are not perfectly comparable to those originally reported; we therefore reproduce Diffusion Policy's results under our setup (Table~\ref{tab:bc_results}).

\paragraph{Goal-Conditioned World Modeling.}
All Explorative World Model runs use XM-10. We evaluate on the single-task Maze2D setting from Table 1 of the Diffuser paper~\cite{janner2022planning}, training for 1M steps instead of 2M and using a goal radius of 0.45. As with Behavior Cloning, for a fair setup, we reproduce Diffuser's results under our setup (Table~\ref{tab:gcwm_results}). Our end-to-end explorative world models' NFE exceeds one only with a recurrent block~\cite{geiping2026scaling}: U-Maze recurs over the full depth 3 times (NFE 4), Large over the middle 3 times (NFE 1.9), Medium not at all.

\paragraph{Language Modeling.}
Our MDLM~\cite{sahoo2024simple} experiments use a context length of 256 and pretrain for 210k steps with a batch size of 64. Models are xxs-sized: 6 transformer blocks, 6 attention heads, and an embedding dimension of 384. We report generative perplexity using the Qwen2.5-1.5B base model (not post-trained)~\cite{yang2024qwen25} as the evaluator, as it is a much better oracle than the commonly used GPT-2~\cite{radford2019language}, though GPT-2 results confirmed the same trends. Results are primarily across an entropy range of 5.0 to 5.7 and the sampling step counts shown in Figure~\ref{fig:mdlm_xm_scaling}. We also confirmed that exploration's gains hold at larger scale as well as for infilling tasks.

%% file: supp_sec/E_related_works.tex
\section{Related Works}
\label{sec:related_work}

\subsection{End-to-End Generative Modeling}
We call a generative model end-to-end when sampling during training, if done at all, is the same as sampling during inference (Section~\ref{sec:e2e_gen_modeling}). Contrastive generative models---including GANs~\cite{goodfellow2014generative} and contrastive-divergence EBMs~\cite{hinton2002training, du2019implicit}---have long been end-to-end, but have struggled with scaling as well as reconstructive models~\cite{dhariwal2021diffusion}, so we focus on reconstructive generative models as the primary target for end-to-end generation. Among reconstructive models, normalizing flows~\cite{dinh2017density} are commonly seen as end-to-end, as they train an invertible map by exact likelihood and sample by inverting that map in a single pass. 
However, they fall short of our definition of end-to-end generative models (Section~\ref{sec:e2e_gen_modeling}), as training only ever evaluates the data-to-noise direction, so the noise-to-data sampling pass used at inference is never simulated during training.
Additionally, normalizing flows have struggled to scale as standalone generators: a continuous bijection cannot split its unimodal base into well-separated modes without leaving thin bridges of density between them~\cite{cornish2020relaxing}. The flows that generate well today escape this by importing factorization, often stacking autoregressive blocks~\cite{zhai2025normalizing, gu2025starflow}, which only widens this train-inference gap, as training never simulates the sequential recursion used at inference. In contrast, recent reconstructive approaches have progressively narrowed the train/inference gap: consistency models~\cite{song2023consistency} learn few-step samplers by enforcing self-consistency along the probability-flow ODE trajectory (via distillation or from scratch), and MeanFlow~\cite{geng2026mean} regresses the flow marginal mean to enable single-step generation. These methods still maintain a mismatch, however, since training primarily samples portions of the trajectory rather than the full generative process used at inference. Mode Forcing explains why this gap has been hard to close: factoring generation into near-unimodal steps is exactly what lets reconstruction losses avoid mode blurring, so existing scalable reconstructive models that factor generation are fundamentally unable to be end-to-end. XMs enable resolving this challenge by factoring the training loop.

\subsection{Coupling}

A natural attempt to avoid mode blurring without factorizing generation is to choose a smarter coupling. Optimal Transport~\cite{pooladian2023multisample, tong2023improving} reduces path crossings by minimizing geometric distance, and minibatch OT~\cite{fatras2021minibatch} approximates this per batch---but the resulting coupling is a biased proxy for the global one, and its gains narrow at scale~\cite{davtyan2025faster}. In our own experiments they reverse entirely (Section~\ref{sec:additional_experiments}), as trying to use OT couplings actually results in worse performance, which we attribute to minibatch bias and OT's exogenous, model-agnostic assignment. Model-Aligned Coupling (MAC)~\cite{lin2026beyond} selects training pairs by model prediction error in addition to geometry, improving few-step flow matching, but is specific to trajectory regularization and not designed to raise generative expressivity or handle multimodal distributions more broadly.

\subsection{Explorative Modeling Based Methods}

We do not claim to invent best-of-$K$: generating several candidates and keeping the best is a simple idea that has appeared across many works, from winner-take-all training objectives~\cite{guzman2012multiple, lee2016stochastic, rupprecht2017learning, li2018implicit, vahabpour2024diverse} to inference-time best-of-$N$ selection, where candidate generations are reranked using a learned reward model~\cite{stiennon2020learning}, a trained verifier~\cite{ma2025inference}, or the model's own energy~\cite{gladstone2025energy}. Rather, what we claim is the realization of why this idea matters for generative modeling: exploration is a fundamentally different way of handling multimodal distributions from the generation factoring of modern reconstructive generative models (described in Section~\ref{sec:background}), working as a scalable latent variable search that keeps the loss minimizer on individual modes rather than their blurred average (Section~\ref{sec:intuition}). In particular, exploration offers a general way to increase generative expressivity, which existing models otherwise fix at design time through how they factor generation. It is this understanding that lets us use exploration deliberately, as a new scaling axis that raises the generative expressivity of existing generative models, and as a standalone approach for end-to-end generation through Forward and Reverse XM. Through this lens, several existing methods can be seen as realizing a form of exploration, which we describe next.

Multiple Choice Learning~\cite{guzman2012multiple, lee2016stochastic} trains an ensemble of $K$ predictors under an oracle loss, gating gradients through whichever member has the lowest error on each example, an idea later extended to single networks with multiple prediction heads~\cite{rupprecht2017learning}. The goal is ensemble diversity for downstream oracle selection, not generative modeling, though it shares the structural idea of backpropagating only through the minimum-loss prediction.

Implicit Maximum Likelihood Estimation (IMLE)~\cite{li2018implicit} draws a pool of model samples, matches each data point to its nearest sample via fast approximate nearest-neighbor search, and trains on those pairs, with theoretical guarantees that this recovers MLE under mild conditions. Structurally, XMs generalize IMLE: IMLE is a specific instance of end-to-end Forward XM with a shared global sample pool in place of per-step candidates, and both minimize the expected distance from each data point to its nearest model output. Self-Organizing Generative Models (SOG)~\cite{vahabpour2024diverse} realize the same objective conditionally---sampling several latent codes per datapoint and training only through the lowest-loss one, another instance of end-to-end Forward XM---interpreting this procedure as encoder-free hard-assignment maximum likelihood. IMLE attributes its success to performing \textit{implicit} maximum likelihood---the argument that nearest-sample matching is a likelihood-free proxy for log-likelihood maximization. But we argue this perspective misidentifies the working mechanism---if maximum likelihood were what makes IMLE work, then Reverse XM, which fixes one model output and trains on its nearest data point rather than the reverse, optimizing the mode-seeking reverse KL rather than any likelihood (Section~\ref{sec:xm_kl_theory}), would have no reason to capture modes---yet it does. What the two methods share instead is exploration: drawing multiple candidates so the reconstruction loss minimizer lands on a mode rather than a blurred average. This coincides with the main hypothesis behind Mode Forcing~\cite{gladstone2026mode}, that modern reconstructive generative models are not fundamentally about MLE or similar objectives but rather about designing a reconstructive objective where the loss minimizer does not blur modes. Explorative Modeling makes this principle general---it applies to any generative model and loss, not just end-to-end implicit models with Euclidean distance, covers both forward and reverse directions, and can be layered onto existing generative models as a scaling axis. 

%% file: supp_sec/F_Additional_Theory.tex
\section{Additional Theory}
\label{sec:additional_theory}

\subsection{Generative Expressivity Details}
\label{sec:generative_expressivity_details}

Here we expand on the definition of generative expressivity $E$ from Section~\ref{sec:mode_forcing}, which follows Mode Forcing~\cite{gladstone2026mode}. The mode count $M(q)$ is the number of strict local maxima of a distribution $q$, or for discrete distributions the cardinality of its support, and $P_\theta(\cdot \mid c)$ is the distribution induced by the model's inference-time sampling procedure. Generative expressivity is
$$
E \;\triangleq\; \sup_{p^*,\, c}\;\; \sup_{\theta^\star \in \arg\min_\theta \mathcal{L}(\theta)} \; M\big(P_{\theta^\star}(\cdot \mid c)\big),
$$
where the outer supremum over data distributions makes $E$ a property of the training objective itself rather than of any particular dataset, the inner supremum reads $E$ as a capacity---the ceiling the objective permits, analogous to parameter count---and minimizers range over all densities (we assume the nonparametric optimum is realizable). Under direct Bregman regression the minimizer is unique for every $p^*$ (the conditional mean, a point mass), so $E = 1$ regardless of parameter count.
Under smooth Forward XM with $K$ candidates, $E \ge K$: taking $p^*$ with $K$ modes separated as in Proposition~3, every minimizer covers each mode with its own region of mass and so retains at least $K$ modes, one maximum per mode (Proposition~3); the separation condition is what makes the mode neighborhoods resolvable.
Achievement is a per-distribution statement: on a $p^*$ with $M^*(c) \le K$ such modes, minimizers retain all $M^*(c)$ of them, so unimodal data yields unimodal minimizers at any $K$. Finally, these claims belong to the smooth objective---the hard min differs by at most $\log K$ at finite $K$, but its large-$K$ limit is a coverage objective without distributional control (Section~\ref{sec:xm_kl_theory}), so there $E$ reports unbounded capacity without implying sample quality. We otherwise inherit the regularity assumptions of Mode Forcing~\cite{gladstone2026mode}---densities exist, reconstruction losses are Bregman divergences, and $M$ counts modes irrespective of their mass or separation---and refer to~\cite{gladstone2026mode} for the mean-blurring lemma behind $E=1$.

\paragraph{Scope of $E$.}
$E$ is an at-optimum capacity: for direct regression it is pinned exactly ($E=1$), and for smooth Forward XM Proposition~3 delivers a lower bound with per-distribution achievement. Applied to \emph{consistent} factored objectives, however, this at-optimum reading saturates. Autoregression under cross-entropy and continuous-time diffusion sampled exactly admit nonparametric minimizers whose sampling distribution is $p^*$ itself (under standard regularity), so at their optima they inherit every mode of the data and $E$ reaches the most the sample space allows---$V^{L}$ for autoregression over $L$ tokens with vocabulary size $V$, unbounded for continuous diffusion (read at the loss scale this count is finite~\cite{gladstone2026mode}, and what a trained model realizes in practice is far smaller, small enough to bind). Nor do finite sampling steps restore a meaningful cap: even a two-step sampler's optimum can retain arbitrarily many well-separated modes, albeit with no control over their mass---whereas Proposition~3 at least guarantees every mode is covered. The at-optimum definition therefore registers no deficit for any consistent objective, because at the optimum there is none---it separates direct regression from factored models, but not factored models from one another.

\paragraph{Per-Prediction Expressivity.}
The deficit factored models do carry lives at the level of each prediction. A factored model is trained as many small reconstruction problems, and each carries its own ceiling---a \emph{per-prediction expressivity}, how much of its own target's structure a single step's minimizer can retain. For an MSE step this ceiling is one mode, the conditional mean of its target; a parallel discrete decoder retains full per-position conditionals but predicts them independently, so no single step retains dependence among the tokens it reveals. Factoring does not raise these per-prediction ceilings; what it changes is each prediction's \emph{residual multimodality}---how many valid targets compete for it during training---which richer conditioning and finer factoring shrink but never quite remove (Section~\ref{sec:mode_forcing}). Mode Forcing makes this picture quantitative, defining each prediction's ceiling $e_t$ and its residual multimodality $M_t(c_t)$ pointwise in the conditioning, so a prediction pays $\max(M_t(c_t) - e_t,\, 0)$ wherever it is asked; these shortfalls are not claimed to sum to a total~\cite{gladstone2026mode}. This residue costs a model in one of two ways. Where sampling exposes the compromise directly it surfaces in generations, as blur or incoherence: direct regression, few jumps, coarse steps, or many positions unmasked at once. Where sampling does not, the same compromise is paid during training instead: competing targets pull each prediction against itself, whether the loss blurs them or must spread over them, a cost set by the training objective and untouched by the sampler. Relieving this training-time cost, we hypothesize, is why exploration can still help models whose sampling steps have already converged (Figures~\ref{fig:xm_explore_fid_fvd_scaling} and~\ref{fig:mdlm_xm_scaling}) and why XMs converge faster (Figure~\ref{fig:xm_rae_flop_step_eff}). So when we say existing models fix generative expressivity at design time, we mean the per-prediction notion: for a given data distribution the factorization sets the residue each prediction faces, the objective sets what each retains of it, and neither changes with parameters or data. Exploration attacks the residue directly, either by letting one input carry $K$ distinct predictions through a latent (end-to-end XMs and our XMDLMs), or by searching which noise draw a datapoint trains under so that fewer competing targets land on the same prediction (our continuous hybrids; Section~\ref{sec:approach_details}). In the budget's terms, the first route raises $e_t$---to at least $K$ for a squared-error prediction by Proposition~3, and heuristically for richer heads, where the $K$ latent-conditioned candidates turn the step's product head into a mixture that can carry dependence among the positions it reveals---while the second shrinks the residue $M_t$ each prediction effectively faces. It should therefore help most where the residue is largest---direct regression, few-jump Jumpy models (the jumps-for-exploration substitution of Figure~\ref{fig:xm_end_to_end_scaling}), and the sparse, any-order conditioning of masked language models. It should help least for autoregression over discrete tokens, whose full-prefix conditioning leaves each next-token target nearly unimodal and whose per-token cross-entropy retains a full conditional rather than a mean, consistent with autoregressive language models proving harder to improve (Section~\ref{sec:conclusion}). Continuous autoregression, where a plain per-step regression would fit each next element by the mean of its valid values, should benefit like the other continuous families. Even a small residue does not make exploration worthless, however: our video generation results improved significantly despite weak multimodality, which we hypothesize reflects surplus generative expressivity aiding optimization and generalization much as overparametrization does (Section~\ref{sec:discussion}).

\subsection{What Forward and Reverse XM Optimize}
\label{sec:xm_kl_theory}

\paragraph{Overview.} This subsection makes the claims of Section~\ref{sec:forward_reverse_xm} precise, and involves two distributions: the data distribution $p^*$ and the distribution of the model's generations $g_\theta$. Because squared error is, up to a constant, the negative log of a Gaussian of width $\sigma$, the loss effectively blurs whichever distribution it is applied to, so we write $p_\theta$ for the model's generations blurred by this Gaussian and $p^*_\sigma$ for the data blurred in the same way. In their smooth forms, Forward and Reverse XM minimize
\[
\mathrm{KL}(p^* \,\|\, p_\theta) + H(p^*)
\qquad\text{and}\qquad
\mathrm{KL}(g_\theta \,\|\, p^*_\sigma) + H(g_\theta),
\]
respectively. These identities hold at every $K$ with the blurred densities replaced by their $K$-sample versions (Propositions 1 and 2), and hold exactly as written in the large-$K$ limit. The two objectives differ only in the entropy each carries. For Forward XM, this entropy is the \emph{data}'s, a constant the model cannot change, so Forward XM performs maximum likelihood---of its $K$-candidate mixture---at every $K$. For Reverse XM, this entropy is instead the \emph{model}'s own, which the model can lower by shrinking its spread, so Reverse XM drifts toward collapse by itself and needs an entropy bonus to become the pure reverse KL. The remainder of this subsection states and proves these claims.

\paragraph{Setup.} We ignore additive constants throughout, as they change neither gradients nor minimizers. We also suppress any conditioning (such as a class label or timestep), as every statement below holds per condition. The loss between a generation $\hat y$ and a datapoint $x$ is then $-\log k_\sigma(\hat y, x)$, where for squared error the kernel is a Gaussian centered on the generation, $k_\sigma(\hat y, x) = \mathcal{N}(x;\, \hat y,\, \sigma^2 I)$. Blurring each side by this kernel gives the two densities of the overview,
\[
p_\theta = g_\theta * k_\sigma, \qquad p^*_\sigma = p^* * k_\sigma,
\]
so that $p_\theta$ represents the model as the loss sees it, and $p^*_\sigma$ represents the data in the same way. Finally, we analyze the smooth form of exploration, which scores the $K$ explored candidates by their average kernel, $-\log \tfrac{1}{K}\sum_i k_\sigma(\hat y_i, x)$, rather than by the best one alone. This soft min differs from the hard min by at most $\log K$, and we describe where the two diverge later in this subsection.

\paragraph{Proposition 1 (Forward XM is maximum likelihood).} \textit{At every $K$, the smooth Forward XM objective (the soft counterpart of Equation~\ref{eq:forward})}
\[
L_F(\theta) = \mathbb{E}_{x \sim p^*}\, \mathbb{E}_{\hat y_{1:K} \sim g_\theta}\!\left[ -\log \tfrac{1}{K}\textstyle\sum_i k_\sigma(\hat y_i, x) \right]
\]
\textit{is exactly the expected negative log-likelihood of the data under the mixture of the model's own $K$ explored generations, $\tfrac{1}{K}\sum_i k_\sigma(\hat y_i, \cdot)$, so Forward XM performs maximum likelihood at every $K$. As $K \to \infty$ this mixture converges to the blurred model $p_\theta$, so minimizing $L_F$ becomes minimizing $\mathrm{KL}(p^* \,\|\, p_\theta)$, recovering $p^*$ as $\sigma \to 0$.}

\begin{proof}[Proof sketch]
For a fixed draw of candidates, $\tfrac{1}{K}\sum_i k_\sigma(\hat y_i, \cdot)$ is a normalized density, so the inner term is exactly the negative log-likelihood of $x$ under this mixture, and subtracting the constant data entropy $H(p^*)$ leaves the forward KL to the mixture. The same average is an unbiased estimate of $p_\theta(x)$, so by Jensen $L_F$ upper-bounds $\mathbb{E}_{p^*}[-\log p_\theta]$ for every $K$, tightening as $K$ grows by the argument of IWAE~\cite{burda2015importance} and converging by the law of large numbers (with domination), where $\mathbb{E}_{p^*}[-\log p_\theta] = \mathrm{KL}(p^* \,\|\, p_\theta) + H(p^*)$.
\end{proof}

\paragraph{Remark (the optimum at fixed $\sigma$).} Over all generator densities, the optimum is the KL projection of $p^*$ onto the blurred family $\{g * k_\sigma\}$; it equals $p^*$ exactly when the $\sigma$-deconvolution of $p^*$ exists as a density, and in general the match becomes exact only as $\sigma \to 0$.

\paragraph{The role of $K$.} Since Forward XM performs maximum likelihood at every $K$, what $K$ controls is the density each draw fits. At $K{=}1$ this density is a single Gaussian of width $\sigma$, whose best fit (for squared error) is a point mass at the data mean, the blurred mean. At larger $K$ it is a $K$-component mixture that can place mass on many modes, becoming the full blurred model $p_\theta$ as $K \to \infty$. At inference, however, the model draws a single sample from $g_\theta$, which the loss scores only through its blur $p_\theta$; from that fixed density's view, $L_F$ is an upper bound on its negative log-likelihood that tightens monotonically as $K$ grows~\cite{burda2015importance}. The gap between the two views is a Jensen gap: because the mixture's components are sampled from $g_\theta$ rather than placed freely, finite $K$ penalizes draws that miss a mode. When per-draw scores concentrate this gap is $O(1/K)$ and merely favors slightly lower-variance generators; with well-separated modes and few candidates, misses are costly enough that minimizers hedge mass between modes (Proposition~3), an effect that fades as $K$ outgrows the mode count. Exploration therefore turns a unimodal regressor into a multimodal likelihood model.

\paragraph{The hard min.} The hard min actually implemented differs from the soft one by at most $\log K$, and the difference matters most in the limit. As $K \to \infty$, the hard objective only asks that every datapoint have some generation arbitrarily close to it, so any model whose generations cover the data's support becomes optimal, regardless of how it spreads mass over that support. The hard min is therefore a \emph{coverage} objective, which shares the mode-covering character of the forward KL without pinning down the density, and the clean likelihood statements are those of the smooth form above.

\paragraph{Proposition 2 (Reverse XM targets the reverse KL, but collapses without an entropy term).} \textit{At every $K$, the smooth Reverse XM objective (the soft counterpart of Equation~\ref{eq:reverse_xm})}
\[
L_R(\theta) = \mathbb{E}_{\hat y \sim g_\theta}\, \mathbb{E}_{x_{1:K} \sim p^*}\!\left[ -\log \tfrac{1}{K}\textstyle\sum_i k_\sigma(\hat y, x_i) \right]
\]
\textit{equals, in expectation over the sampled targets, $\mathrm{KL}(g_\theta \,\|\, \hat p_\sigma) + H(g_\theta)$, where $\hat p_\sigma = \tfrac{1}{K}\sum_i k_\sigma(\cdot, x_i)$ is the kernel density estimate of the $K$ data targets, converging to $\mathrm{KL}(g_\theta \,\|\, p^*_\sigma) + H(g_\theta)$ as $K \to \infty$. Because each generation is scored on its own, independently of how often the model produces it, $L_R$ is linear in $g_\theta$, so its infimum is approached by collapse onto a point mass (at $K{=}1$, the data mean) and bare Reverse XM does not recover $p^*$. Adding an entropy bonus, i.e.\ minimizing $L_R - H(g_\theta)$, cancels the entropy term and leaves the pure reverse KL, minimized as $K \to \infty$ at $g_\theta = p^*_\sigma$, reaching $p^*$ as $\sigma \to 0$.}

\begin{proof}[Proof sketch]
For a fixed draw of targets, $\hat p_\sigma$ is a normalized density, so the inner expectation is the cross-entropy between $g_\theta$ and $\hat p_\sigma$, which decomposes as $\mathrm{KL}(g_\theta \,\|\, \hat p_\sigma) + H(g_\theta)$; by the law of large numbers $\hat p_\sigma(\hat y) \to p^*_\sigma(\hat y)$, giving the limit. The objective weights each generation by a score that does not depend on how often the model produces it, so it is linear in $g_\theta$, and its infimum over densities is approached by densities concentrating toward a point mass at the minimizer of the per-generation cost (the data mean at $K{=}1$, a global mode of $p^*_\sigma$ as $K$ grows). Subtracting $H(g_\theta)$ in the limit leaves the reverse KL, minimized at $g_\theta = p^*_\sigma$.
\end{proof}

\paragraph{Where the blur sits.} One further asymmetry is where the blur sits. Forward XM compares the raw data against the blurred model, whereas Reverse XM compares the raw model against the blurred data, and as $\sigma \to 0$ both become comparisons between the raw model and the raw data, in opposite directions.

\paragraph{On the ELBO.} Only the Forward side carries a genuine evidence bound, as the inner $\log \tfrac{1}{K}\sum_i k_\sigma(\hat y_i, x)$ is, in expectation, an importance-weighted lower bound on the model log-likelihood $\log p_\theta(x)$, as in IWAE~\cite{burda2015importance}. The entropy-corrected Reverse objective $L_R - H(g_\theta)$ is instead a variational free energy for the target $p^*_\sigma$ as $K \to \infty$, while bare $L_R$ keeps only its energy term, which is why it collapses and bounds no model likelihood. For this reason, we describe Reverse XM as reverse-KL or variational rather than as an ELBO.

\paragraph{Assumptions.} These statements require (i) the loss to be, up to an additive constant, the negative log of a kernel whose normalizer is independent of $\hat y$ and $\theta$ (squared error $\leftrightarrow$ Gaussian; for non-normalizable losses such as bounded or perceptual ones the reading fails and these statements do not apply), (ii) the smooth surrogate rather than the hard min, (iii) $K \to \infty$ only to replace the $K$-sample mixtures with the blurred densities $p_\theta$ and $p^*_\sigma$, and $\sigma \to 0$ to reach $p^*$ exactly, and (iv) mild regularity: a bounded, symmetric, translation-invariant kernel (so $p^*_\sigma$ is a density), finite expected loss and differential entropies, domination to justify the limit interchanges, a continuous loss vanishing only at $\hat y = x$ (for the hard-min limit), and minimizers ranging over all densities. Finally, these statements analyze end-to-end XMs, but because conditioning is suppressed throughout, they extend to any single prediction of a factored model whose $K$ explored candidates share their conditioning and target, with $p^*$ read as that prediction's target conditional. For discrete predictions the kernel is the candidate's own softmax, so Proposition~1's mixture-likelihood reading carries over with no blur ($\sigma$ plays no role), which covers our XMDLMs; the separation-based Proposition~3 and the kernel-density reading of Proposition~2 are squared-error statements and do not transfer. The noise-searching form of our continuous hybrids instead pairs each candidate with its own corruption of the datapoint, a coupling search rather than a best-of-$K$ against a fixed target, and we leave its analysis, along with how per-prediction gains compose across a sampler's steps, to future work.

\paragraph{Proposition 3 (Expressivity of smooth Forward XM under separation).} \textit{Suppose $p^*$ is a mixture of $M^* \le K$ unimodal components with near-equal masses, each concentrated at scale $\sigma$ or below, whose modes are pairwise separated by distances $\Delta \gg \sigma$, with $\Delta^2/\sigma^2 \gg (\log K)/\min_m w_m$ for component masses $w_m$, and let the assumptions of this section hold. Then every minimizer $g_\theta$ of the smooth Forward XM objective $L_F$ places positive mass in a neighborhood of each mode of $p^*$---so $M(g_\theta) \ge M^*$, and taking $M^* = K$ gives $E \ge K$---though at finite $K$ minimizers may also hedge mass between modes, as insurance against candidate draws that miss a mode. Once candidates sufficiently outnumber modes, $K(1 - \min_m w_m)^K\, \Delta^2/\sigma^2 \lesssim 1$, misses are rare enough that hedging no longer pays: all but a negligible fraction of mass sits at the modes, and the blurred density $p_\theta = g_\theta * k_\sigma$ has exactly $M^*$ strict local maxima, one per component---equivalently, what the model deploys, $g_\theta$, carries exactly $M^*$ modes read at the loss scale.}

\begin{proof}[Proof sketch]
By Proposition~1, minimizing $L_F$ is maximum likelihood of $p^*$ under the mixture of $K$ candidates drawn i.i.d.\ from $g_\theta$. If $g_\theta$ assigns a mode zero mass, no draw ever covers it, and the objective then pays of order $w_m \Delta^2/\sigma^2$ for a mode of mass $w_m$, while concentrating that mass elsewhere gains at most $\log K$ in likelihood, plus what it saves by making misses rarer at the remaining modes---with near-equal masses a constant fraction of what the dropped mode costs---so neither term covers the loss, and every minimizer covers every mode with positive mass. The same $\log K$ budget also forces concentration: spreading mode $m$'s share to width $\rho$ costs its covered datapoints of order $w_m\rho^2/\sigma^2$ against a total possible gain of $\log K$, so under the separation condition each mode's mass sits within $o(\Delta)$ of its center. Each region's peak density therefore dominates the between-mode bridges, whose height is capped by the same budget, and hedge mass---lying at distance $\sim\Delta$ from the peaks at lower density---leaves each peak a strict local maximum, hence $M(g_\theta)\ge M^*$; concentration to $O(\sigma)$ and the exact count $M^*$ require the rare-miss condition below.
Because candidates are sampled rather than placed, a draw misses a mode of weight $w$ with probability $(1-w)^K$, at a cost of order $\Delta^2/\sigma^2$, so minimizers also keep insurance mass \emph{between} modes---the bridges visible at small $K$ in Figure~\ref{fig:exploration_improves_coverage}---which likewise perturbs the optimal mode weights. Enumerating $K$ learned discrete latents instead puts a candidate at every mode on every step, so misses never occur and the bridges vanish at any $K$, though $K$ latents alone then cap how many distinct outputs a condition can produce (it's possible a mixture of discrete and continuous is optimal).
Under the additional condition, misses are rare enough that insurance costs more in diluted likelihood than it saves, so all but a negligible fraction of mass concentrates in the mode neighborhoods (the concentration hypothesis leaves exponentially little of $p^*$ elsewhere). Convolving with $k_\sigma$ then merges each neighborhood's sub-$\sigma$ structure into a single bump, with exponentially small cross-terms and no spurious maxima between components, so $p_\theta$ has exactly $M^*$ strict local maxima.
\end{proof}

%% file: supp_sec/G_FAQ.tex
\section{Frequently Asked Questions}
\label{sec:faq}

Here we answer some common questions about Explorative Modeling.

\paragraph{What's the main takeaway? How is this paper `novel' if it's just best-of-$K$?}
Introducing best-of-$K$ is not the central contribution or claim of this paper, as sampling candidates and keeping the best has appeared many times before~\cite{lee2016stochastic, li2018implicit, vahabpour2024diverse} (Section~\ref{sec:related_work}). What is new is the realization of what this simple loop does: it increases generative expressivity without factoring generation, which is a completely different factorization axis from popular generative models such as Autoregression and Diffusion (Figure~\ref{fig:gen_modeling_axes}). The paper's central messages follow from this realization: generative expressivity is worth scaling through exploration alongside model parameters and data as a new pretraining axis, as it improves both the performance and generalization of existing generative models (a way of trading training compute for generalization). Because generative expressivity increasingly becomes the bottleneck as parameters and data grow, gains from exploration \textbf{grow with scale} rather than saturate, making exploration more important as scale increases. Exploration can also stand in for factoring generation, turning how end-to-end a model is into something we can scale rather than a fixed design choice, all the way to fully end-to-end XMs that match diffusion baselines at a fraction of the inference compute. Finally, best-of-$K$ is just one implementation of exploration---we have already seen other implementations succeed (Reverse XM, Section~\ref{sec:additional_experiments}), and expect more versions to eventually become viable (e.g., gradient-based search, Section~\ref{sec:future_work}).

\paragraph{Does exploration make inference more expensive?}
No. Exploration happens entirely during training, so inference is unchanged. The added cost shows up as more expensive training instead (for Forward XM, for Reverse XM the added training cost is often negligible), and as shown in Figure~\ref{subfig:xrae_flop_eff}, this cost is well worth it.

\paragraph{Is exploration really a new ``axis'' if it just costs more compute?}
Scaling parameters or data costs compute too---no scaling axis is free. The question is whether exploration is a \textit{good} way to spend compute, which is the same question compute-optimal scaling~\cite{hoffmann2022training} asks when dividing a budget between parameters and data. Our FLOP-matched comparisons answer it directly: models that explore are significantly more compute efficient than models that just train for longer (Figures~\ref{subfig:xrae_flop_eff} and~\ref{subfig:fid_xm_flops_scaling}), and the FLOP-optimal amount of exploration grows with scale, just as the compute-optimal parameter count does.

\paragraph{Why the name Explorative Modeling?}
We believe the idea of exploration captures the intuition of trying to understand the model's loss landscape with respect to the different modes. However, exploration is a type of search, and we considered naming the approach after search directly. We decided against this because gradient descent is already a search over parameters, and because search is now strongly associated with inference, where ``scaling search'' would sound like inference-time thinking/reasoning rather than a training axis. Every candidate term is overloaded in some way; we found exploration to be the best general term, covering Forward XM, Reverse XM, and the gradient-based variants discussed in Section~\ref{sec:future_work}.

\paragraph{When does exploration help?}
Exploration helps whenever generation is conditioned on a latent variable the model can search over, such as the input noise in a diffusion model. We have found exploration helps most when generative expressivity is a bottleneck, although this does not have to be the case---our video models improved significantly even though the generated distribution was not very multimodal (Section~\ref{sec:discussion}).

\paragraph{Does training on the model's own generations cause collapse?}
No, because the training targets are always real data: the model's generations only decide which datapoint each noise input gets paired with. In Forward XM, every datapoint still receives a training signal, so no part of the data distribution can be dropped; in fact, Forward XM is maximum likelihood (Section~\ref{sec:xm_kl_theory}). Collapse is a genuine concern for Reverse XM, which is mode-seeking, and is exactly why it needs an entropy term. A simple coverage constraint served as this entropy term when training our Reverse XM language models (Section~\ref{sec:additional_experiments}).

\paragraph{How should the amount of exploration $K$ be chosen?}
This deserves more study, just like studying how parameters and data should be scaled together has been important~\cite{hoffmann2022training}. The right amount of exploration is problem dependent, as some distributions have many more modes to capture and so rely on exploration more than others. In our experiments, higher $K$ values became better with scale, both within a single run, where the FLOP-optimal amount of exploration grows over the course of training (Figures~\ref{subfig:xrae_flop_eff} and~\ref{subfig:fid_xm_flops_scaling}), and across runs, where gains from exploration grow with model and data scale (Figure~\ref{fig:xm_data_param_gains_scale}). Exploration is also flexible: models with more training compute available can increase exploration to buy generalization and more end-to-end generation. Given these many tradeoffs, we recommend sweeping $K$ across $1$ (the baseline), $2$, $3$, and $5$ to start, sweeping $8$ and $12$ if compute permits, and pushing even further if gains continue. When compute is tight, we recommend starting with small values like $K=2$ or $3$, as they are cheap and improve performance significantly (Figure~\ref{fig:xm_rae_flop_step_eff}).

\paragraph{Is Explorative Modeling a form of reinforcement learning?}
No. Explorative Modeling is a generative modeling objective, unrelated to reinforcement learning. RL uses ``exploration'' for agents trying varied actions to discover reward~\cite{sutton1998reinforcement}, which differs in mechanism and purpose from XMs' within-step candidate sampling, though both share the intuition that trying many options reveals information a single choice cannot.